\documentclass{article} 

\def\iclrpreprintcopy{}
\usepackage{iclr2027_conference,times}

\usepackage{amsmath,amsfonts,bm}

\def\eqref#1{equation~\ref{#1}}

\def\1{\bm{1}}

\DeclareMathAlphabet{\mathsfit}{\encodingdefault}{\sfdefault}{m}{sl}
\SetMathAlphabet{\mathsfit}{bold}{\encodingdefault}{\sfdefault}{bx}{n}

\usepackage[utf8]{inputenc}
\usepackage[T1]{fontenc}
\usepackage{hyperref}
\usepackage{url}
\usepackage{graphicx}
\usepackage{booktabs}
\usepackage{amsfonts}
\usepackage{amsmath}
\usepackage{nicefrac}
\usepackage{microtype}
\usepackage{xcolor}
\usepackage{subcaption}
\usepackage{multirow}

\newsavebox{\resultstablebox}
\newcommand{\fitresultstable}[1]{%
  \sbox{\resultstablebox}{#1}%
  \ifdim\wd\resultstablebox>\linewidth
    \resizebox{\linewidth}{!}{\usebox{\resultstablebox}}%
  \else
    \usebox{\resultstablebox}%
  \fi}

\graphicspath{{./}{iclr2027/}}

\title{OPERA: Object Perception Enhances\\
Single-view 3D Reconstruction}

\author{Y Huynh$^1$, Duc Thanh Nguyen$^2$, Mohamed Abdelrazek$^1$ \\
        $^1$Applied Artificial Intelligence Initiative, Deakin University \\
        $^2$School of Information Technology, Deakin University}

\begin{document}

\raggedbottom

\maketitle

\begingroup
\setlength{\intextsep}{6pt}
\begin{figure}[!htb]
\centering
\includegraphics[width=0.8\linewidth]{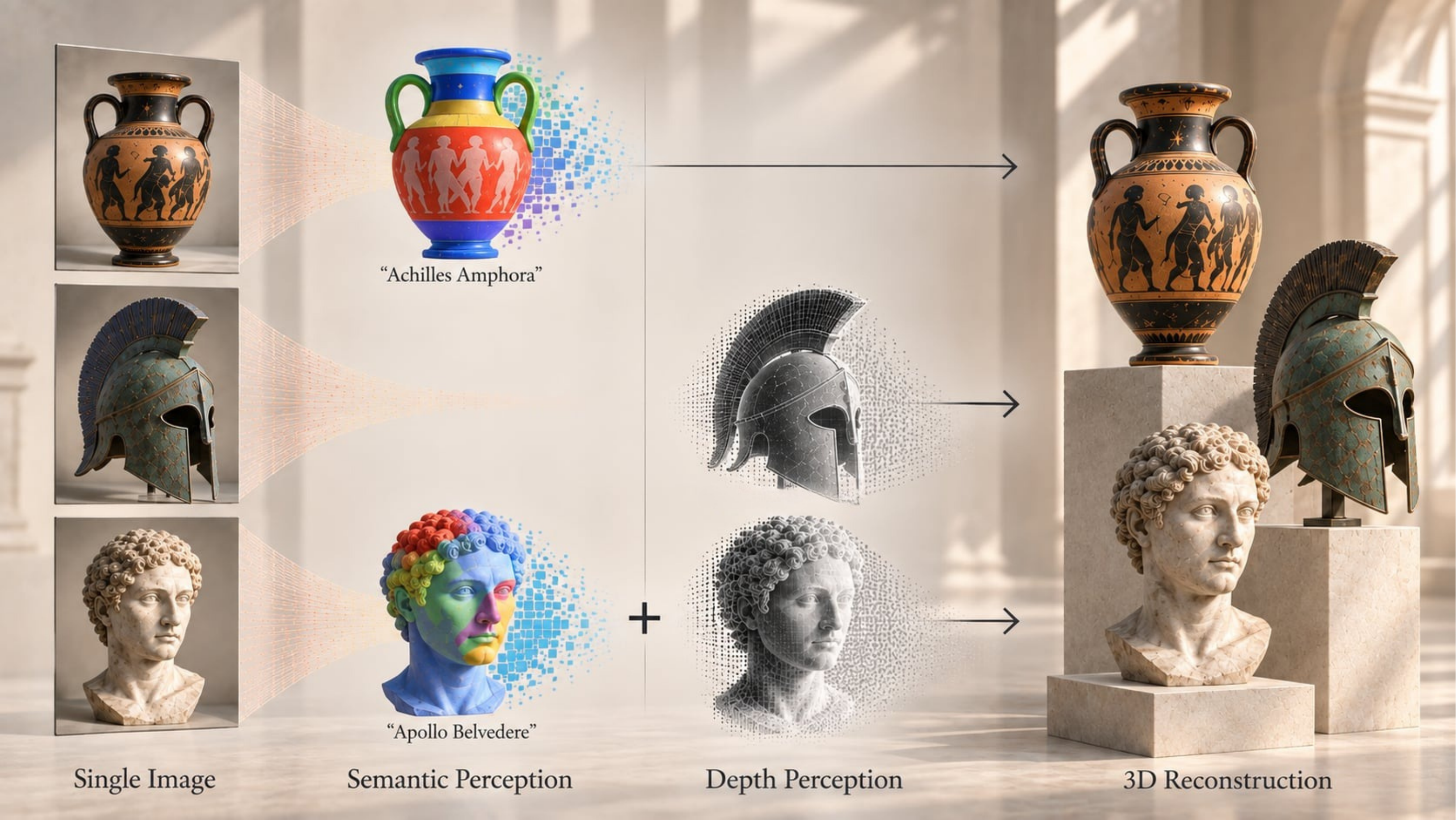}
\caption{\textbf{Perception-guided single-view 3D reconstruction.} Semantic and depth perception guide reconstruction individually or jointly. This illustration does not depict experimental results.}
\label{fig:teaser}
\end{figure}
\endgroup

\begin{abstract}
Single-view 3D reconstruction is a challenging task in computer vision due to information missing from the single input image. Generative model-based approaches can produce plausible 3D objects from a single image, thanks to data-driven priors learnt from rich and large-scale datasets. However, plausible generation does not guarantee fidelity to the geometry and appearance of the particular input object. Inspired by object perception in human vision, we propose OPERA, a framework that guides multi-view diffusion sampling with pretrained perception models through lightweight alignment modules. These modules are trained independently while both the generative and perception models remain frozen, allowing multiple signals to be combined at inference without joint fusion training. We evaluate OPERA on two single-view 3D reconstruction baselines using subsets of Google Scanned Objects and OmniObject3D. On the primary baseline, combined guidance reduces mean Chamfer Distance by 30.4\% and 24.8\%, respectively, relative to unguided reconstruction. We also compare our method with recent image-to-3D models. We provide in-depth analyses of the design choices and their effects across datasets and backbones. Our project page is at \url{https://opera-3d.github.io/}.
\end{abstract}

\section{Introduction}
\label{sec:intro}

Single-view 3D object reconstruction aims to recover the shape and
appearance of the particular object depicted in an image, not merely
generate a plausible object of the same category. This distinction is
fundamentally about \emph{instance fidelity}: preserving the proportions,
surface patterns, and distinctive features of that object. Yet recovering
a complete object from a single observation is inherently ill-posed
because unseen surfaces remain ambiguous. Existing approaches address
this ambiguity through learnt shape
priors~\citep{choy20163d,mescheder2019occupancy} and generative modelling.
Multi-view diffusion pipelines synthesise novel views before reconstructing a 3D
surface~\citep{liu2023syncdreamer,wonder3d,era3d}.
Image-to-3D models such as SAM~3D Objects~\citep{sam3dteam2025sam3d3dfyimages}
and TRELLIS.2~\citep{Xiang_2026_CVPR} generate complete objects from
partial visual evidence. However, generative completion can introduce
hallucinated details that are visually plausible but inconsistent with
the particular object being generated. Such outputs may preserve the
broad category while altering object-defining features, as illustrated
in Figure~\ref{fig:qualitative_3d}. In multi-view pipelines, errors in
generated shape, appearance, or cross-view consistency can propagate
into the reconstruction. Our task remains reconstruction: generation
must serve fidelity to the observed object, rather than plausibility
alone. This motivates using object-specific perceptual knowledge as
explicit guidance to constrain generation towards faithful reconstruction.


Human vision motivates this perspective: object recognition
and prior knowledge contribute to interpreting incomplete
visual observations~\citep{biederman1987recognition,palmer1999vision}.
Pretrained perception models offer a computational means of introducing
object-specific cues into reconstruction. Vision--language models
describe object attributes~\citep{DBLP:journals/corr/abs-2502-13923},
which text encoders represent as semantic
features~\citep{DBLP:journals/jmlr/RaffelSRLNMZLL20}.
Self-supervised visual models provide spatially structured
representations~\citep{DBLP:journals/tmlr/OquabDMVSKFHMEA24,dblp:journals/tmlr/SimeoniVSB0JKSM26},
while monocular depth models infer geometric
cues~\citep{DBLP:conf/nips/YangKH0XFZ24}.
These signals express knowledge acquired through different pretraining
objectives, rather than new observations of unseen surfaces.
This motivates a broader question:
\emph{can independently trained perception models provide useful
constraints on multi-view generation, improving downstream
3D reconstruction without fine-tuning the underlying models?}
Such an approach would allow perception, generation, and
reconstruction to contribute their specialised capabilities
without requiring a jointly trained replacement for the
entire pipeline.

We propose OPERA, a framework that incorporates object perception
into multi-view diffusion sampling. This includes a
generation-perception alignment module (GPAM) that maps intermediate
generative features into each target perceptual space, enabling
alignment objectives to steer multi-view generation.
Each GPAM is trained independently while the generative and perception
models remain frozen, and multiple GPAMs can be combined at inference
without joint fusion training. OPERA augments existing multi-view
diffusion pipelines through alignment modules trained separately
for each backbone and perceptual target.
Figure~\ref{fig:teaser} illustrates this use of semantic and geometric
knowledge to guide reconstruction.

We instantiate the framework with language-derived semantic targets,
spatial visual representations, and depth features, and evaluate it
on Wonder3D~\citep{wonder3d} and Era3D~\citep{era3d} using GSO
and OmniObject3D~\citep{downs_google_2022,DBLP:conf/cvpr/WuZFWRPWYWQLL23}.
In the reported configurations, combined guidance achieves lower mean
Chamfer Distance than either modality alone across both pipelines and
datasets. Analyses of target representation, supervised views, and
guidance strength characterise how these choices affect geometry and
appearance. Together, these results provide a proof of concept for
incorporating independently trained perceptual capabilities into
generative reconstruction through task-specific alignment modules.

We summarise our contributions as follows:
\begin{itemize}
    \item We introduce OPERA, a framework that leverages perceptual signals from
    pretrained perception models to drive multi-view diffusion sampling for single-view 3D reconstruction, without backbone fine-tuning or joint fusion training.
    \item We design GPAMs and target-specific alignment objectives for global semantic vectors and spatial feature maps, supporting
    individual or combined guidance and view-dependent alignment for spatial targets.

    \item We evaluate OPERA on two single-view 3D reconstruction pipelines and two benchmark subsets, and provide in-depth analyses of how perceptual targets and guidance configurations affect geometry and appearance.
\end{itemize}

\section{Related Work}
\label{sec:related}

\subsection{3D Object Reconstruction}
3D object reconstruction aims to recover an object's geometry from one
or more images. Classical methods use correspondence and
triangulation~\citep{Schonberger_2016_CVPR}; neural methods optimise
representations against observed images, including
NeRF~\citep{mildenhall2020nerf}, 3D Gaussian
Splatting~\citep{kerbl20233dgs}, and surface-based
NeuS~\citep{DBLP:conf/nips/WangLLTKW21}.
With fewer observations, learnt priors become important:
3D-R2N2~\citep{choy20163d} predicts volumetric shape from one or
more images, while Occupancy Networks~\citep{mescheder2019occupancy}
predict a continuous occupancy field.

Single-view reconstruction relies particularly on learnt shape
priors~\citep{DBLP:conf/cvpr/TatarchenkoRRLK19} to resolve missing
observations. Diffusion models~\citep{DBLP:conf/cvpr/RombachBLEO22}
address this ambiguity through novel-view synthesis:
Zero-1-to-3~\citep{liu2023zero123} introduces view-conditioned diffusion,
and SyncDreamer~\citep{liu2023syncdreamer} jointly generates consistent
views. Wonder3D~\citep{wonder3d} generates colour and normal views
through cross-domain diffusion; Era3D~\citep{era3d} adds camera
prediction and efficient row-wise attention. These views support
surface reconstruction, while InstantMesh~\citep{xu2024instantmesh}
and CRM~\citep{wang2024crm} pair multi-view generation with feed-forward
reconstruction. Image-to-3D models such as SAM~3D
Objects~\citep{sam3dteam2025sam3d3dfyimages} and
TRELLIS.2~\citep{Xiang_2026_CVPR} instead generate 3D representations
without an explicit intermediate set of novel-view images.
Sampling-time control can use classifier
gradients~\citep{DBLP:conf/nips/DhariwalN21} or predictions from
intermediate diffusion features~\citep{DBLP:conf/cvpr/LuoDWGH24}.
OPERA adapts feature-based guidance to coupled multi-view, cross-domain
generation through target-specific global and spatial objectives.
Rather than training a replacement generator, we evaluate whether
input-derived perceptual constraints improve downstream reconstruction.

\subsection{Perception for 3D Reconstruction}
Perceptual cues complement geometric evidence, particularly when
observations are limited. Panoptic 3D reconstruction jointly predicts
geometry, semantic labels, and object instances from a single RGB
image~\citep{DBLP:conf/nips/DahnertHND21}.
Other methods incorporate perceptual priors across views:
MS-GS combines depth priors and local semantic regions with
virtual-view supervision~\citep{li2025msgs}, while PE3R integrates
multi-view geometry and 2D semantic priors for tuning-free semantic
reconstruction~\citep{DBLP:journals/corr/abs-2503-07507}.
These approaches complement geometric evidence with information about
\emph{what} is observed, through object and region identification,
depth priors, and semantic constraints.

Our distinction lies in where perceptual knowledge enters the pipeline.
Rather than using perception directly for geometric or semantic
reconstruction, OPERA uses input-derived perceptual objectives to guide
multi-view generation before surface reconstruction. GPAMs map
intermediate generative features into the target spaces of pretrained
perception models, accommodating global semantic vectors and spatial
feature maps. Each GPAM is trained for its backbone and perceptual
target while the generative and perception models remain frozen.
This design allows independently trained perceptual signals to be
combined at inference without backbone fine-tuning or joint fusion
training, while retaining the existing downstream reconstruction stage.

\section{Methodology}
\label{sec:methodology}

\subsection{Overview}
\label{sec:overview}

\begin{figure}[ht]
    \centering
    \includegraphics[width=1.0\linewidth]{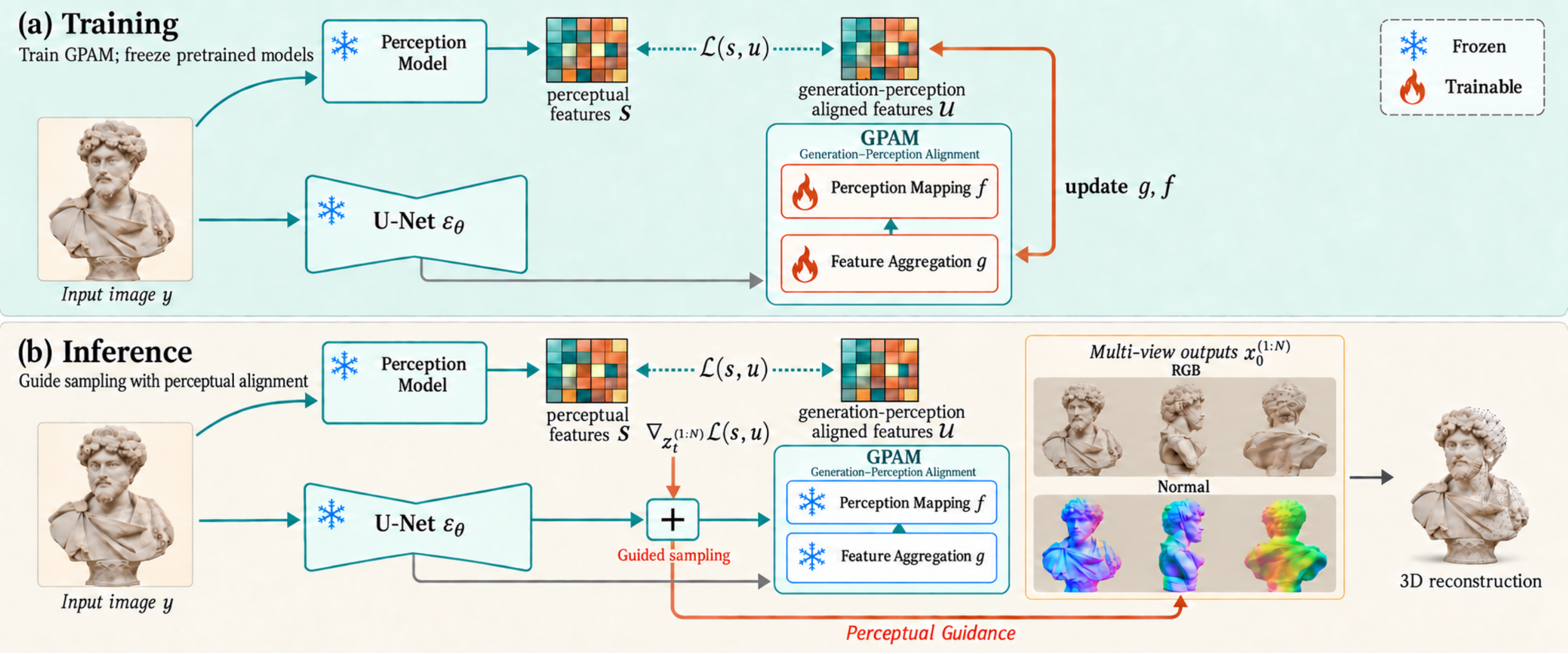}
    \caption{\textbf{Overview of OPERA.} Frozen perception models provide targets for independently trained GPAMs. At inference, alignment gradients update the multi-view latents before the frozen U-Net is re-evaluated for denoising.}
    \label{fig:framework}
\end{figure}

Figure~\ref{fig:framework} illustrates our framework for reconstructing
$\mathbf{X}$ from its image $\mathbf{y}$.
Multi-view diffusion pipelines~\citep{liu2023syncdreamer,wonder3d,era3d}
condition on $\mathbf{y}$ to generate $N$ views in a pretrained
autoencoder's latent space~\citep{DBLP:conf/cvpr/RombachBLEO22}.
The sampling state $\mathbf{z}^{(1:N)}_t$ evolves from Gaussian noise;
the input-image encoding supplies conditioning.
Decoded views $\mathbf{x}^{(1:N)}_{t=0}$ support surface
reconstruction~\citep{DBLP:conf/nips/WangLLTKW21}.
The background DDPM reverse-process
parameterisation~\citep{DBLP:conf/nips/HoJA20} is
\begin{equation}
\begin{aligned}
    &p_{\theta}\big(\mathbf{z}_{0:T}^{(1:N)}|\mathbf{y}\big)=p\big(\mathbf{z}_{T}^{(1:N)}|\mathbf{y}\big)\prod_{t=1}^T\prod_{n=1}^N p_{\theta}\big(\mathbf{z}_{t-1}^{(n)}|\mathbf{z}_t^{(1:N)},\mathbf{y}\big) \\
    &p_{\theta}\big(\mathbf{z}_{t-1}^{(n)}|\mathbf{z}_t^{(1:N)},\mathbf{y}\big) = \mathcal{N}\big( \mathbf{z}_{t-1}^{(n)};\mu_{\theta}^{(n)}(\mathbf{z}_{t}^{(1:N)},\mathbf{y},t),\sigma_t^2 \mathbf{I} \big) \\
    &\mu_{\theta}^{(n)}\big(\mathbf{z}_{t}^{(1:N)},\mathbf{y},t\big)=\frac{1}{\sqrt{\alpha_t}} \bigg( \mathbf{z}_t^{(n)} - \frac{\beta_t}{\sqrt{1-\bar{\alpha}_t}} \epsilon_{\theta}^{(n)}\big(\mathbf{z}_t^{(1:N)},\mathbf{y},t \big) \bigg)
\end{aligned}
\label{eq:multi_view_diffusion}
\end{equation}
where $\beta_t$ is the noise schedule, $\alpha_t=1-\beta_t$, $\bar{\alpha}_t=\prod_{t'=1}^t \alpha_{t'}$, and $\sigma_t$ is a prescribed transition standard deviation. The pretrained U-Net $\epsilon_{\theta}$ predicts noise for all $N$ views from their joint state and $\mathbf{y}$. This background parameterisation is distinct from the DDIM sampling updates used below.

The generated normal maps provide geometric supervision for surface
reconstruction, while colour images supervise appearance.

We use object-specific perceptual knowledge to guide the multi-view
sampling. This knowledge enters through respective GPAMs without backbone fine-tuning or additional conditioning. Specifically, let $s$ denote the perceptual features that a frozen perception model extracts from $\mathbf{y}$, and let $u$ denote the generation-perception aligned features that GPAM predicts from the intermediate activations of the frozen U-Net $\epsilon_{\theta}$. 


Classifier guidance~\citep{DBLP:conf/nips/DhariwalN21} steers a diffusion model by correcting the predicted noise using the gradient of an auxiliary classifier,
\begin{align}
    \epsilon_t^{(1:N)} \leftarrow \epsilon_{\theta}^{(1:N)}\big( \mathbf{z}_t^{(1:N)},\mathbf{y},t \big) - \sqrt{1-\bar{\alpha}_t}\,\nabla_{\mathbf{z}_t^{(1:N)}} \log p_{\phi}\big(\ell \,\big|\, \mathbf{z}_t^{(1:N)}\big),
    \label{eq:classifier_guidance}
\end{align}
where $\ell$ is the desired condition and $p_{\phi}$ is a time-aware
classifier trained on noisy inputs. OPERA instead differentiates the
GPAM loss with respect to the multi-view latents, updates them, and
re-evaluates the frozen U-Net before the DDIM
transition~\citep{DBLP:conf/iclr/SongME21}, rather than
directly correcting its noise prediction. For $K$ modalities with
losses $\mathcal{L}_k(s_k,u_k)$, combined guidance uses
\begin{equation}
\mathbf{z}_{t}^{(1:N)}
\leftarrow
\mathbf{z}_{t}^{(1:N)}
-
\sum_{k=1}^{K}
\lambda_k \eta_k\,
\mathcal{R}\!\left(
\nabla_{\mathbf{z}_{t}^{(1:N)}}
\mathcal{L}_k(s_k,u_k)
\right),
\label{eq:multi_modality_update}
\end{equation}
where $\eta_k$ is the guidance scale and $\lambda_k$ the mixing weight
of modality $k$, with $\sum_{k=1}^{K}\lambda_k=1$.
The operator $\mathcal{R}$ normalises each modality's gradient to unit
RMS within each colour or normal domain before mixing, so
$\lambda_k\eta_k$ sets its update scale independently of the raw
gradient magnitude. The state $\mathbf{z}^{(1:N)}_t$ includes both
domains; only their latents are updated, not network parameters.
Single-modality update rules are given in Appendix~\ref{app:latent_updates}.

Each modality-specific GPAM is trained independently against its perception target. Adding a modality requires its own GPAM, objective, and gradient evaluation, without retraining existing modules or the backbone. GPAM parameters are backbone-specific, while the guidance formulation is reused across the two evaluated backbones (Section~\ref{sec:setup}). During guided sampling, perceptual feature extraction retains the computational graph so that perception losses can be differentiated through GPAMs and the U-Net $\epsilon_{\theta}$ with respect to the latents, while all network parameters remain frozen.

\subsection{Perception Modalities}
\label{sec:perception_modalities}

We demonstrate our approach with two types of perceptual knowledge: semantic information via
language and visual descriptions, and geometric information via depth sensing.

\paragraph{Semantic perception.}
Semantic targets describe object identity, texture, and appearance.
We investigate language-derived and visual representations to capture the semantics. For language-derived perception, we use Qwen2.5-VL~\citep{DBLP:journals/corr/abs-2502-13923} to generate a caption for the input image $\mathbf{y}$, then encode the caption with T5~\citep{DBLP:journals/jmlr/RaffelSRLNMZLL20}. This results in a global semantic representation $s_{\mathrm{cap}}\in\mathbb{R}^{D_{\mathrm{cap}}}$. For visual perception, we extract patch-based features from $\mathbf{y}$ using 
DINOv2~\citep{DBLP:journals/tmlr/OquabDMVSKFHMEA24} or
DINOv3~\citep{dblp:journals/tmlr/SimeoniVSB0JKSM26}, then arrange these patch-based features into a feature map
$s_{\mathrm{vis}}\in
\mathbb{R}^{D_{\mathrm{vis}}\times H_{\mathrm{vis}}\times W_{\mathrm{vis}}}$
capturing both local appearance and spatial layout.

\paragraph{Geometric perception.}
Monocular depth estimation provides geometric cues about the depicted
object. We extract intermediate depth features from
DepthAnything-V2~\citep{DBLP:conf/nips/YangKH0XFZ24} to form
$s_{\mathrm{depth}}\in
\mathbb{R}^{D_{\mathrm{depth}}\times H_{\mathrm{depth}}\times W_{\mathrm{depth}}}$.
Using features rather than final depth maps is motivated by
perceptual-loss approaches~\citep{DBLP:conf/eccv/JohnsonAF16};
Appendix~\ref{app:depth_ablation} examines this choice empirically.

\subsection{Generation-Perception Alignment}
\label{sec:perception_alignment}

Each perceptual target has its own GPAM, which maps intermediate 
activations in the U-Net $\epsilon_{\theta}$ into the target space to define a differentiable alignment
objective for latent guidance~\citep{DBLP:conf/nips/DhariwalN21,DBLP:conf/cvpr/LuoDWGH24}.
For coupled multi-view, cross-domain generation, GPAM comprises two
sub-modules: \textbf{Feature Aggregation}, which combines backbone-specific
activations into a compact multi-view representation, and
\textbf{Perception Mapping}, which projects this representation into the
target space while preserving its global or spatial structure. Backbone-specific architectures
and implementation details are provided in Appendix~\ref{app:architecture}.

\paragraph{Feature Aggregation.}
We extract intermediate activations from selected decoder stages of
the frozen U-Net $\epsilon_{\theta}$. Each stage is processed
independently to obtain a common channel dimension and spatial
resolution before the features are combined. The processing blocks and
order of resizing are backbone-specific (Appendix~\ref{app:feature_aggregation}). The processed features are combined using learned per-stage weights
normalised by a softmax. The resulting representation preserves
spatial layout and keeps views separate, supporting global or
spatial predictions.

\paragraph{Perception Mapping.} Depending on the perception modality used, a modality-specific perception mapping module is designed. In principle, perception mapping aims to project the aggregated features, achieved from the feature aggregation module, into the
target perceptual space. 

\emph{Global mapping.}
For global semantic representations, i.e., Qwen2.5-VL + T5 or called caption-T5 hereafter for brevity, we spatially pool the aggregated features and
average the resulting descriptors over the views supplied to the
GPAM. A two-layer MLP with a GELU activation maps this descriptor
to the aligned feature $u$. This pooling matches the object-level
structure of the language-derived target without imposing
correspondences between text features and image locations.

\emph{Spatial mapping.}
For spatial DINO and depth targets, we retain the spatial layout of
the aggregated colour and normal features. A lightweight convolutional
decoder predicts aligned maps for the selected front, front-left, and
front-right views. The target shares the front-view pose; neighbouring
predictions are compared structurally, not pointwise.



\subsection{Training and Inference}
\label{sec:training_inference}

Each GPAM is trained independently for its perception target and
generative backbone, while the backbone and perception model remain
frozen. Specifically, generative features are extracted from the forward-noised
input-image latents replicated across view slots from the U-Net model $\epsilon_{\theta}$. The GPAM learns to map the generative features with the input image's perceptual target and is subsequently
applied to evolving multi-view sampling states at inference. The loss $\mathcal{L}(s,u)$ is defined accordingly.

\paragraph{Global targets.}
For caption--T5, we define the loss $\mathcal{L}_{\text{global}}(s,u)$ using the cosine distance between the object-level perceptual features $s$
and GPAM's learnt features $u$:
\begin{equation}
    \mathcal{L}_{\mathrm{global}}(s,u)
    = 1 - \cos(s,u).
    \label{eq:semantic_loss}
\end{equation}


\paragraph{Spatial targets.}
For spatial DINO and depth targets, we match the front-view prediction
$u^{(0)}$ directly to the input-derived target $s$ and compare each
auxiliary-view prediction structurally to the same target:
\begin{equation}
    \mathcal{L}_{\mathrm{spatial}}(s,u)
    =
    \ell_{2\text{-}\mathrm{std}}(s,u^{(0)})
    +
    \sum_{v\in\mathcal{A}}
    \omega_v\,\ell_{\mathrm{CKA}}(s,u^{(v)}).
    \label{eq:geometric_loss}
\end{equation}
where $\ell_{2\text{-}\mathrm{std}}$ is MSE between standardised maps
and $\ell_{\mathrm{CKA}}=1-\mathrm{CKA}$ uses linear centred kernel
alignment~\citep{DBLP:conf/icml/Kornblith0LH19}.
The auxiliary set $\mathcal{A}$ excludes the front view and contains
front-left and front-right in the main configuration.
Weights $\omega_v=1/|\mathcal{A}|$ are therefore $1/2$ each, preserving
the total auxiliary weight when coverage changes.


\paragraph{Inference.}
The same objective structure supplies the latent gradients in
Eq.~(\ref{eq:multi_modality_update}).
The target is computed once per input image, while predictions
are recomputed from the current denoising state.
Training and inference differ in the standardisation used for
depth features; Appendix~\ref{app:objective_details} specifies the
normalisation axes and the pooled DINO variant.

\section{Experimental Setup}
\label{sec:setup}
\label{sec:experimental_setup}


\paragraph{Implementation details.}
We evaluate OPERA with Wonder3D~\citep{wonder3d} and
Era3D~\citep{era3d}. For each backbone and perception target, a respective GPAM is trained. We set $\eta_d=2.5\times10^{-3}$ for depth guidance,
$\eta_s=10^{-3}$ for semantic guidance, and $(\lambda_d,\lambda_s)=(0.75,0.25)$ for combined
guidance (Eq.~(\ref{eq:multi_modality_update})).
For our primary backbone, Era3D, sampling uses 40 DDIM steps with
a classifier-free guidance scale of 3.0, unchanged from its baseline.
All experiments use an NVIDIA H100 80G.
Multi-view generation takes approximately 2.5 minutes per object,
followed by approximately 2 minutes for NeuS reconstruction.
See more details in Appendix~\ref{app:objective_details}. Perception models and parameter settings were tuned with Era3D, and then applied to Wonder3D.

\paragraph{Datasets.}
We train GPAMs on 1{,}000 objects from
Objaverse~\citep{DBLP:conf/cvpr/DeitkeSSWMVSEKF23}, split into
900 training and 100 validation objects.
Evaluation uses 30 objects from Google Scanned Objects
(GSO)~\citep{downs_google_2022} and 100 from
OmniObject3D~\citep{DBLP:conf/cvpr/WuZFWRPWYWQLL23}.

\paragraph{Evaluation metrics.}
We report six metrics: Chamfer Distance (CD$\downarrow$) between
meshes normalised to a unit sphere; volumetric IoU (IoU$\uparrow$)
on a $256^3$ grid; F1@0.1$\uparrow$, with the distance threshold
expressed in unit-sphere radii; and PSNR$\uparrow$, SSIM$\uparrow$,
and LPIPS$\downarrow$ on novel views against rendered
ground truth. Performance tables report means over 30 GSO objects or 100
OmniObject3D objects, with CD in units of $10^{-3}$.
Bold marks the best mean per metric within each comparison group.

\section{Main Results}
\label{sec:results}
\label{sec:main_results}

Tables~\ref{tab:main_era3d} and~\ref{tab:main_wonder3d} compare each
backbone with semantic, depth, and combined guidance.

\paragraph{Quantitative results.}
Combined guidance improves all six mean metrics over the baseline
on both datasets and backbones. On GSO, it also achieves the best
mean value on every metric among the four settings for each backbone.
On OmniObject3D, it has lower mean CD than either single modality,
but depth alone retains higher IoU on Era3D and higher F1@0.1 on
Wonder3D. The magnitude of improvement varies: the relative reductions of mean CD are
30.4\% and 24.8\% for Era3D--GSO and Era3D--OmniObject3D, respectively,
and 6.5\% and 0.5\% for the corresponding Wonder3D evaluations.
On Wonder3D, the single modalities yield smaller mean CD changes than
their combination; semantic guidance alone slightly increases CD on GSO
(18.17 versus 18.14). These mean-only comparisons do not establish
whether small differences are reproducible across sampling seeds.
These results support configuration reuse across backbones.
Appendix~\ref{app:paired_statistics} reports paired object-level
analyses, and Appendix~\ref{app:external_comparison} provides further
comparisons with modern image-to-3D models (SAM~3D
Objects~\citep{sam3dteam2025sam3d3dfyimages} and
TRELLIS.2~\citep{Xiang_2026_CVPR}).

\begin{table}[!htbp]
\centering
\begin{minipage}[t]{0.49\linewidth}
\vspace{0pt}
\centering
\captionsetup{font=small}
\scriptsize
\setlength{\tabcolsep}{1.5pt}
\renewcommand{\arraystretch}{0.95}
\caption{Quantitative evaluation on Era3D.}
\label{tab:main_era3d}
\fitresultstable{%
\begin{tabular}{@{}lrrrrrr@{}}
\toprule
Setting & CD$\downarrow$ & IoU$\uparrow$ & F1@.1$\uparrow$ & PSNR$\uparrow$ & SSIM$\uparrow$ & LPIPS$\downarrow$ \\
\midrule
\multicolumn{7}{l}{\textit{GSO}} \\
Baseline & 16.46 & 0.531 & 0.801 & 21.54 & 0.9222 & 0.0705 \\
+ Depth & 11.64 & 0.558 & 0.833 & 22.21 & 0.9272 & 0.0624 \\
+ Semantic & 15.51 & 0.529 & 0.803 & 21.70 & 0.9235 & 0.0690 \\
+ Combined & \textbf{11.45} & \textbf{0.574} & \textbf{0.842} & \textbf{22.42} & \textbf{0.9283} & \textbf{0.0613} \\
\midrule
\multicolumn{7}{l}{\textit{OmniObject3D}} \\
Baseline & 35.78 & 0.499 & 0.683 & 20.44 & 0.9162 & 0.0842 \\
+ Depth & 33.49 & \textbf{0.526} & 0.698 & 20.71 & 0.9179 & 0.0804 \\
+ Semantic & 28.89 & 0.503 & 0.693 & 21.02 & 0.9183 & 0.0773 \\
+ Combined & \textbf{26.89} & 0.512 & \textbf{0.705} & \textbf{21.25} & \textbf{0.9205} & \textbf{0.0746} \\
\bottomrule
\end{tabular}%
}
\end{minipage}%
\hfill
\begin{minipage}[t]{0.49\linewidth}
\vspace{0pt}
\centering
\captionsetup{font=small}
\scriptsize
\setlength{\tabcolsep}{1.5pt}
\renewcommand{\arraystretch}{0.95}
\caption{Quantitative evaluation on Wonder3D.}
\label{tab:main_wonder3d}
\fitresultstable{%
\begin{tabular}{@{}lrrrrrr@{}}
\toprule
Setting & CD$\downarrow$ & IoU$\uparrow$ & F1@.1$\uparrow$ & PSNR$\uparrow$ & SSIM$\uparrow$ & LPIPS$\downarrow$ \\
\midrule
\multicolumn{7}{l}{\textit{GSO}} \\
Baseline & 18.14 & 0.544 & 0.798 & 21.72 & 0.9249 & 0.0694 \\
+ Depth & 17.76 & 0.547 & 0.800 & 21.77 & 0.9252 & 0.0690 \\
+ Semantic & 18.17 & 0.544 & 0.801 & 21.74 & 0.9252 & 0.0691 \\
+ Combined & \textbf{16.96} & \textbf{0.554} & \textbf{0.814} & \textbf{21.93} & \textbf{0.9263} & \textbf{0.0671} \\
\midrule
\multicolumn{7}{l}{\textit{OmniObject3D}} \\
Baseline & 21.68 & 0.577 & 0.764 & 22.38 & 0.9317 & 0.0670 \\
+ Depth & 21.68 & 0.590 & \textbf{0.778} & 22.52 & 0.9327 & 0.0660 \\
+ Semantic & 21.65 & 0.582 & 0.770 & 22.45 & 0.9322 & 0.0664 \\
+ Combined & \textbf{21.57} & \textbf{0.595} & 0.776 & \textbf{22.73} & \textbf{0.9335} & \textbf{0.0649} \\
\bottomrule
\end{tabular}%
}
\end{minipage}
\par\smallskip
{\footnotesize
CD is reported in units of $10^{-3}$. Caption--T5 and DINOv3 are used as semantic models on GSO and OmniObject3D, respectively.
\par}
\end{table}

\begin{figure}[ht]
    \centering
    \includegraphics[width=1.0\linewidth]{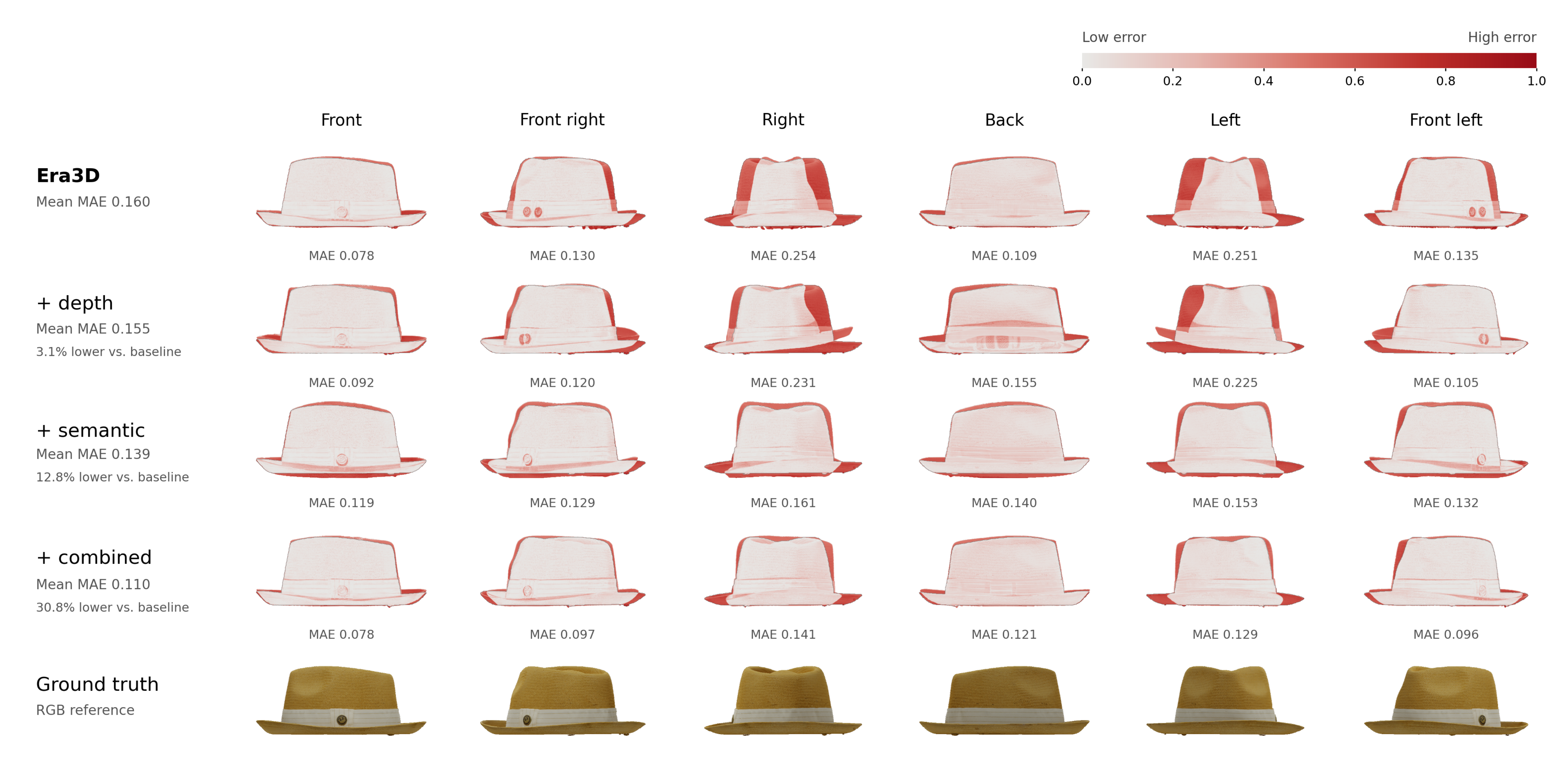}
    \caption{\textbf{Effects of perceptual guidance on multi-view generation.}
    Error maps of the directly generated RGB views relative to ground truth
    for one object, comparing Era3D with depth, semantic, and combined guidance. Stronger
    red indicates larger RGB error. Foreground MAE is reported per view
    and averaged across the six views.}
    \label{fig:modality_effect_2d}
\end{figure}

\paragraph{Qualitative results.}
Figure~\ref{fig:modality_effect_2d} compares foreground RGB errors in
Era3D's six generated views before reconstruction. For this object,
depth and semantic guidance reduce mean absolute error (MAE) from
0.160 to 0.155 and 0.139, respectively; combined guidance reduces it
further to 0.110. Gains are view-dependent: both single modalities
increase front-view error, while combined guidance increases back-view
error relative to the baseline. This example illustrates potential
complementarity, not uniform improvement.


\begin{figure}[t]
    \centering
    \includegraphics[width=0.9\linewidth,height=0.78\textheight,keepaspectratio]{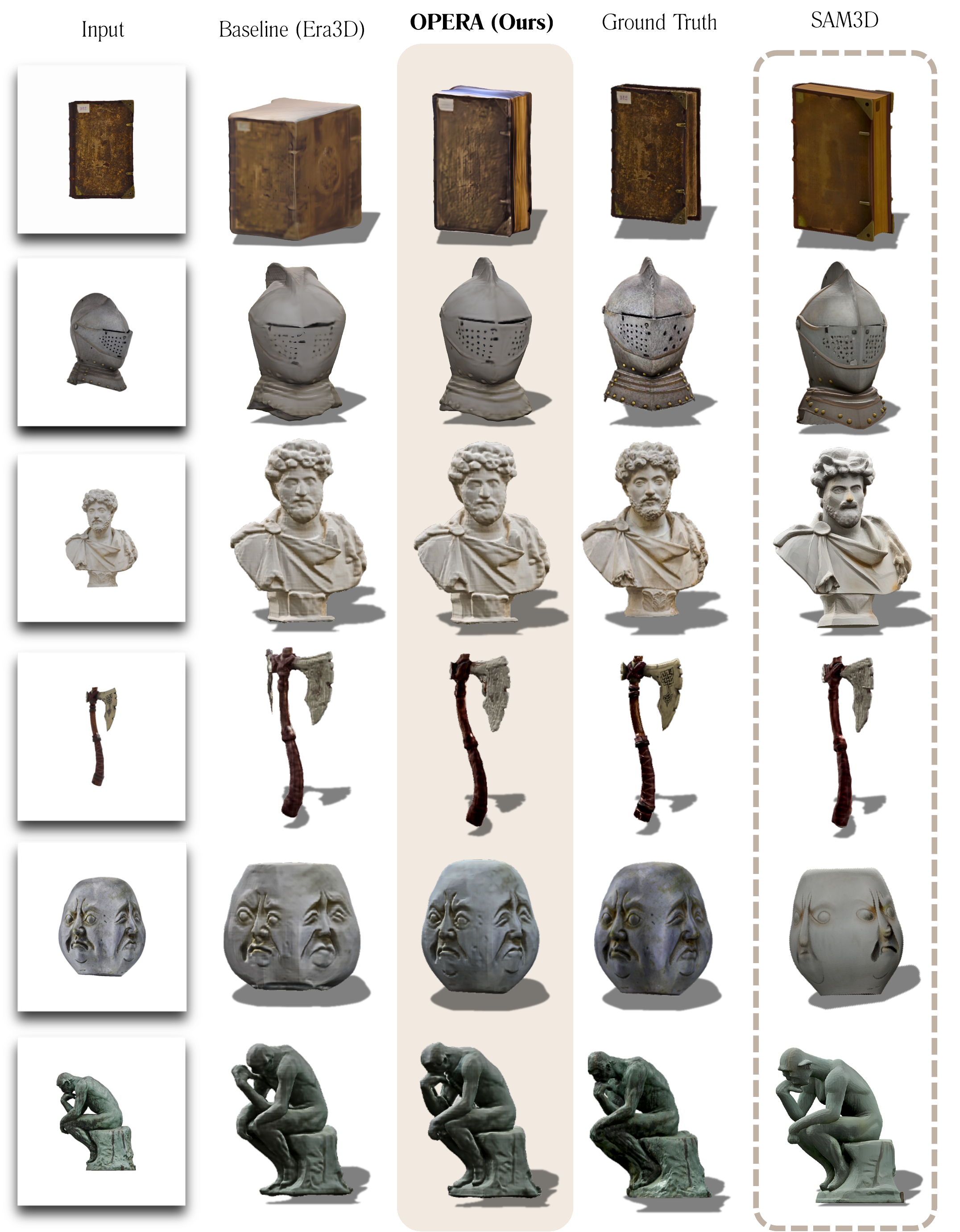}
    \caption{\textbf{Perception-guided 3D reconstruction with Era3D.}
    From left to right: input image, Era3D baseline, Era3D with OPERA,
    and ground truth. SAM~3D Objects~\citep{sam3dteam2025sam3d3dfyimages}
    is shown separately as an external image-to-3D reference.}
    \label{fig:qualitative_3d}
    \label{fig:qualitative}
\end{figure}

\paragraph{Effects on 3D reconstruction.}
Figure~\ref{fig:qualitative_3d} illustrates changes in object proportions
and surface appearance, including reduced book bulk after guidance.
SAM~3D Objects is included as an external image-to-3D reference.
In the displayed examples, its outputs show reduced facial relief
on the four-faced sculpture, altered facial features on the
Marcus bust, and more angular body contours on \emph{The Thinker}.
These contrasts illustrate that a plausible overall shape does not
necessarily preserve an object's distinctive appearance. They concern
selected examples rather than a general ranking; details already
present in Era3D should not be attributed to OPERA.
Appendix~\ref{app:external_comparison} provides further quantitative
and qualitative comparisons with external image-to-3D generators,
examining both geometric accuracy and object-specific appearance fidelity.


\section{Ablation Studies}
\label{sec:analysis}
\label{sec:ablations}

We examine our method under different model selections for each modality
and modality combinations, using Era3D as the baseline for these
ablation studies. The comparisons motivate a shared configuration
rather than a separate optimum for each metric.
Additional experiments on semantic and depth perception, including
guidance scales and depth representation selection, are
provided in Appendices~\ref{app:semantic_ablation}
and~\ref{app:depth_ablation}.

\paragraph{Semantic target selection across datasets.}
Table~\ref{tab:semantic_targets} compares the three semantic targets
at the shared guidance scale $\eta_s=10^{-3}$.
Their relative performance varies across datasets and metrics. On GSO, Caption--T5 achieves the
highest F1@0.1 and best appearance metrics, although DINOv2 has lower
CD and DINOv3 higher IoU.
On OmniObject3D, DINOv3 has the lowest CD and best appearance metrics,
while DINOv2 has higher IoU and F1@0.1.
We therefore select Caption--T5 for GSO and DINOv3 for OmniObject3D
based on their geometry and appearance results, rather than superiority
on every metric.
Spatial DINOv3 is a starting candidate for broader object
collections, subject to validation on the target backbone.

\begin{table}[!htbp]
\centering
\small
\caption{Semantic target comparison on Era3D.}
\label{tab:semantic_targets}
\begin{minipage}[t]{0.49\linewidth}
\vspace{0pt}
\centering
\scriptsize
\setlength{\tabcolsep}{1.5pt}
\renewcommand{\arraystretch}{0.95}
\textbf{GSO}\par\smallskip
\fitresultstable{%
\begin{tabular}{@{}lrrrrrr@{}}
\toprule
Target & CD$\downarrow$ & IoU$\uparrow$ & F1@.1$\uparrow$ & PSNR$\uparrow$ & SSIM$\uparrow$ & LPIPS$\downarrow$ \\
\midrule
Caption--T5 & 15.51 & 0.529 & \textbf{0.803} & \textbf{21.70} & \textbf{0.9235} & \textbf{0.0690} \\
DINOv2 & \textbf{15.13} & 0.534 & 0.800 & 21.43 & 0.9213 & 0.0695 \\
DINOv3 & 16.28 & \textbf{0.538} & 0.796 & 21.31 & 0.9203 & 0.0714 \\
\bottomrule
\end{tabular}%
}
\end{minipage}%
\hfill
\begin{minipage}[t]{0.49\linewidth}
\vspace{0pt}
\centering
\scriptsize
\setlength{\tabcolsep}{1.5pt}
\renewcommand{\arraystretch}{0.95}
\textbf{OmniObject3D}\par\smallskip
\fitresultstable{%
\begin{tabular}{@{}lrrrrrr@{}}
\toprule
Target & CD$\downarrow$ & IoU$\uparrow$ & F1@.1$\uparrow$ & PSNR$\uparrow$ & SSIM$\uparrow$ & LPIPS$\downarrow$ \\
\midrule
Caption--T5 & 34.59 & 0.503 & 0.681 & 20.43 & 0.9163 & 0.0842 \\
DINOv2 & 29.97 & \textbf{0.505} & \textbf{0.694} & 20.95 & 0.9182 & 0.0775 \\
DINOv3 & \textbf{28.89} & 0.503 & 0.693 & \textbf{21.02} & \textbf{0.9183} & \textbf{0.0773} \\
\bottomrule
\end{tabular}%
}
\end{minipage}
\end{table}

\paragraph{Depth representation and guidance scale.}
On Era3D--GSO, intermediate depth features with $\eta_d=2.5\times10^{-3}$
give the lowest CD among the tested depth configurations
(Table~\ref{tab:depth_representation}).
For each semantic target, $\eta_s=10^{-3}$ gives the lowest CD
(Table~\ref{tab:learning_rate}).
The complete comparisons are in Appendices~\ref{app:depth_representation}
and~\ref{app:guidance_step_size}.

\paragraph{Mixing weights.}
With the selected targets and guidance scales,
$(\lambda_d,\lambda_s)=(0.75,0.25)$ achieves the best mean value
on all six metrics among the tested ratios on both datasets
(Table~\ref{tab:mixing_weights}).
The selected mixture also has lower mean CD than either single modality
on both datasets (Table~\ref{tab:main_era3d}), supporting combination
over either component alone in these evaluations, without establishing
that $\lambda_d=0.75$ is optimal between the tested ratios.
Because $\eta_d\ne\eta_s$, varying the mixing weights changes both
modality balance and the nominal total guidance coefficient
$\lambda_d\eta_d+\lambda_s\eta_s$.

\begin{table}[!htbp]
\centering
\small
\caption{Mixing-weight comparison on Era3D.}
\label{tab:mixing_weights}
\begin{minipage}[t]{0.49\linewidth}
\vspace{0pt}
\centering
\scriptsize
\setlength{\tabcolsep}{1.5pt}
\renewcommand{\arraystretch}{0.95}
\textbf{GSO: depth + Caption--T5}\par\smallskip
\fitresultstable{%
\begin{tabular}{@{}lrrrrrr@{}}
\toprule
$\lambda_d:\lambda_s$ & CD$\downarrow$ & IoU$\uparrow$ & F1@.1$\uparrow$ & PSNR$\uparrow$ & SSIM$\uparrow$ & LPIPS$\downarrow$ \\
\midrule
$0.75:0.25$ & \textbf{11.45} & \textbf{0.574} & \textbf{0.842} & \textbf{22.42} & \textbf{0.9283} & \textbf{0.0613} \\
$0.50:0.50$ & 11.93 & 0.555 & 0.833 & 22.19 & 0.9273 & 0.0628 \\
$0.25:0.75$ & 14.03 & 0.550 & 0.823 & 21.96 & 0.9253 & 0.0659 \\
\bottomrule
\end{tabular}%
}
\end{minipage}%
\hfill
\begin{minipage}[t]{0.49\linewidth}
\vspace{0pt}
\centering
\scriptsize
\setlength{\tabcolsep}{1.5pt}
\renewcommand{\arraystretch}{0.95}
\textbf{OmniObject3D: depth + DINOv3}\par\smallskip
\fitresultstable{%
\begin{tabular}{@{}lrrrrrr@{}}
\toprule
$\lambda_d:\lambda_s$ & CD$\downarrow$ & IoU$\uparrow$ & F1@.1$\uparrow$ & PSNR$\uparrow$ & SSIM$\uparrow$ & LPIPS$\downarrow$ \\
\midrule
$0.75:0.25$ & \textbf{26.89} & \textbf{0.512} & \textbf{0.705} & \textbf{21.25} & \textbf{0.9205} & \textbf{0.0746} \\
$0.50:0.50$ & 27.82 & 0.505 & 0.696 & 21.16 & 0.9197 & 0.0754 \\
$0.25:0.75$ & 28.21 & 0.503 & 0.698 & 21.08 & 0.9193 & 0.0759 \\
\bottomrule
\end{tabular}%
}
\end{minipage}
\end{table}

\paragraph{DINOv3 GPAM configuration.}
Spatial DINOv3 has lower CD and better appearance metrics than
the pooled variant, which has higher IoU and F1@0.1
(Table~\ref{tab:gpam_configuration}, Appendix~\ref{app:gpam_configuration}).
This supports our design choice, but does not isolate spatial
information because supervised views and the objective also differ.
Appendix~\ref{app:gpam_configuration} additionally isolates the
inference-time auxiliary CKA terms using the same spatial GPAM.

\section{Discussion and Conclusion}
\label{sec:discussion_conclusion}
\label{sec:limitations}
\label{sec:conclusion}
\paragraph{Limitations and Future Work.}
OPERA has been evaluated within multi-view diffusion pipelines.
Extending perceptual guidance to flow-based 3D generators requires
alignment modules and update rules adapted to their representations
and sampling dynamics. Inference remains costly: our pipeline takes around 4 minutes per
object, including around 2 minutes for downstream NeuS reconstruction.
Guidance adds gradient computation and denoiser evaluations during
sampling. Although modular training avoids joint retraining of the
underlying models, it does not eliminate inference overhead.
Selective guidance schedules and faster reconstruction stages could
improve efficiency.

\paragraph{Conclusion.}

We presented OPERA, a modular method that uses pretrained object
perception to guide generative 3D reconstruction. Experiments on two
pipelines and two benchmarks show geometry and appearance gains that
vary with the guidance configuration. These results establish a proof
of concept for integrating independently trained perceptual capabilities
without jointly retraining the reconstruction system, motivating
extensions to other generative formulations and variable multi-view
inputs to improve fidelity through learnt priors and additional observations.

\section*{Reproducibility Statement}

Section~\ref{sec:methodology} specifies the perceptual targets, alignment
objectives, and latent update. Section~\ref{sec:setup} describes the data
splits, evaluation metrics, and configuration-selection procedure.
Appendix~\ref{app:architecture} provides implementation details, and
Appendices~\ref{app:semantic_ablation} and~\ref{app:depth_ablation}
report additional configuration comparisons.

\section*{Ethics Statement}

Single-view reconstructions contain inferred geometry and should not
be treated as verified records of unseen surfaces. Perceptual guidance
can also propagate errors and biases from pretrained models. Applications
requiring geometric accuracy should therefore validate reconstructions
against additional observations. Use and redistribution of pretrained
models and benchmark assets remain subject to their respective licences.

\section*{AI Use Statement}
Generative AI tools assisted with manuscript drafting and editing,
literature search, feedback on experimental design and interpretation
of results, and preparation of visual materials.
The authors take full responsibility for the methodology,
experimental results, claims, and references presented in this paper.

\bibliographystyle{iclr2027_conference}
\bibliography{references}

\clearpage
\appendix
\section{GPAM Architecture and Implementation Details}
\label{sec:appendix}
\label{app:architecture}
\begin{figure}[!htbp]
\centering
\includegraphics[width=\linewidth]{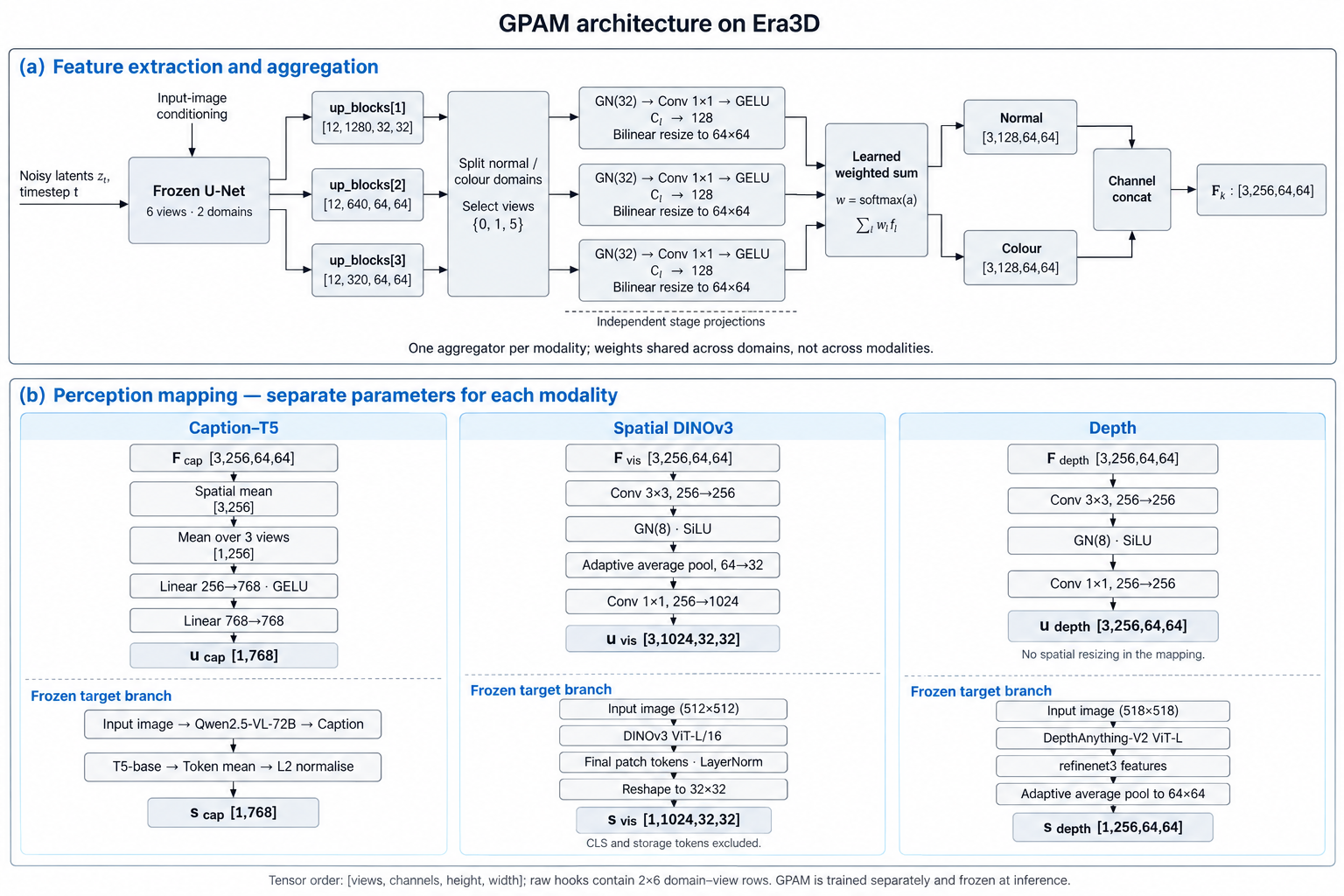}
\caption{\textbf{GPAM architecture on Era3D.}
(a) Decoder activations are separated by domain and restricted to
views $\{0,1,5\}$ before stage-wise projection and learned aggregation.
Normal- and colour-domain features are then concatenated along channels.
(b) Target-specific mappings produce a global Caption--T5 vector or
spatial DINOv3 and depth features $u$, matching the representation
of the corresponding frozen target $s$. The aggregation structure is
shared across GPAMs, but their parameters are separate. Within each
GPAM, aggregation weights are shared across domains, views, and
timesteps. GPAM parameters are trained and then frozen for guidance.}
\label{fig:gpam_architecture}
\end{figure}

Figure~\ref{fig:gpam_architecture} details the Feature Aggregation and
Perception Mapping modules for the Era3D configuration. Each perceptual
target has its own GPAM, trained independently of the other GPAMs and
the frozen backbone. We describe this configuration first, followed
by the architectural adaptations for Wonder3D. Tensor shapes below
are given for one object; four-dimensional feature tensors use the
order [views, channels, height, width].

\subsection{What GPAM Learns}
\label{app:gpam_learning}
\paragraph{Training inputs and supervision.}
GPAM learns to predict an input-derived perceptual target $s$ from
the frozen U-Net's intermediate activations. For Era3D, the front-image
VAE latent is replicated across six view slots and both generated
domains. Each copy receives independent Gaussian noise at a shared,
uniformly sampled training timestep. The frozen U-Net processes these
states conditioned on the input image, and its detached activations
provide the inputs to GPAM. The training pair therefore consists of
features of forward-noised, replicated input-image latents and the
perceptual target extracted from that same image, rather than
ground-truth multi-view observations or states collected from sampling
trajectories. Only the Feature Aggregation and Perception Mapping
parameters are optimised; the generative and perception models remain
frozen.

\paragraph{Meaning of the prediction.}
The aligned feature $u$ is a prediction in the chosen perceptual space,
not a reconstructed image or mesh. For Caption--T5, GPAM predicts a
global caption-derived vector from pooled features. For DINO and depth,
it predicts spatial feature maps, with direct front-view alignment and
auxiliary CKA comparisons against the same input-image target.
The auxiliary predictions are not supervised by target features
extracted from unseen views: CKA provides a relaxed structural objective
rather than pointwise supervision of those views. Consequently,
$\mathcal{L}(s,u)$ measures agreement with the input-derived perceptual
representation, not reconstruction error against ground-truth geometry.

\paragraph{Use during generation.}
At inference, GPAM is applied to evolving multi-view sampling states,
rather than the replicated input-image states used for training.
All network parameters are frozen, but feature extraction retains the
computational graph. Differentiating the alignment loss through GPAM
and the U-Net supplies a direction for updating the latents, after
which the denoiser is re-evaluated before the sampling transition.
Restricting the views used by GPAM does not restrict the latent update
to those views: the coupled U-Net provides gradient paths to all six
views in both generated domains. Reconstruction metrics assess whether
this perceptual guidance improves the downstream result.

\subsection{Feature Aggregation}
\label{app:feature_aggregation}
We register forward hooks on three decoder blocks of the frozen
U-Net, \texttt{up\_blocks[1,2,3]}. For the conditional forward pass
over six views and two domains, their feature shapes are
$[12,1280,32,32]$, $[12,640,64,64]$, and $[12,320,64,64]$,
respectively. We separate the normal and colour domains and select
the front, front-right, and front-left views (indices $0$, $1$, and
$5$) from each stage before aggregation.

Each stage has an independent projection consisting of GroupNorm
with 32 groups, a bias-free $1\times1$ convolution to 128 channels,
and GELU, followed by bilinear resizing to $64\times64$.
These Era3D projection blocks are non-residual. Three learned logits
are normalised by a softmax to weight the processed stage features
before summation. One aggregator is shared across the two domains
within each GPAM, yielding $[3,128,64,64]$ features per domain.
Concatenating the normal and colour features along channels produces
$[3,256,64,64]$ features for Perception Mapping. This operation retains
the spatial layout and keeps the three views separate.

\subsection{Perception Mapping}
\label{app:perception_mapping}
\paragraph{Caption--T5.}
Global average pooling over spatial dimensions produces a
256-dimensional descriptor for each supplied view. Averaging these
descriptors across the three views gives a single object-level vector.
A two-layer MLP, with widths $256\to768\to768$ and GELU between
the linear layers, maps it to the aligned feature $u$ of size
$[1,768]$. Both generated domains contribute to this Era3D prediction.

\paragraph{Spatial DINOv3.}
A lightweight convolutional decoder applies a $3\times3$ convolution
from 256 to 256 channels, GroupNorm with eight groups, and SiLU.
Adaptive average pooling reduces the spatial resolution from
$64\times64$ to $32\times32$, and a $1\times1$ convolution projects
to 1024 channels. The output $u$ has size $[3,1024,32,32]$, preserving
a separate spatial prediction for each supervised view.

\paragraph{Depth.}
The depth decoder uses the same $3\times3$ convolution, GroupNorm,
and SiLU sequence, followed directly by a $1\times1$ convolution
to 256 channels. It requires no spatial resizing and produces $u$
of size $[3,256,64,64]$. The depth and DINOv3 decoders have separate
parameters; their output resolutions match their respective targets.

\subsection{Perceptual Target Extraction}
\label{app:target_extraction}
The target models introduced in Section~\ref{sec:perception_modalities}
remain frozen. Each target is extracted from the input image, not
from the generated neighbouring views.

\paragraph{Caption--T5.}
Qwen2.5-VL-72B-Instruct produces an object description using a fixed
prompt about identity, appearance, shape, and distinctive parts,
with greedy decoding. T5-base encodes the caption; its token embeddings
are mean-pooled and $\ell_2$-normalised to obtain $s\in\mathbb{R}^{768}$.

\paragraph{Spatial DINOv3.}
We use DINOv3 ViT-L/16 pretrained on LVD-1689M. The input is resized
to $512\times512$ and normalised using ImageNet statistics.
The final-block, LayerNorm-normalised patch tokens are reshaped into
a $[1,1024,32,32]$ target. The class token and four storage tokens
are excluded; no spatial pooling is applied to this target.

\paragraph{Depth.}
DepthAnything-V2 ViT-L receives the input resized to $518\times518$
and normalised using ImageNet statistics. We extract intermediate
depth features from its \texttt{refinenet3} refinement layer and
adaptively average-pool them to obtain $s$ of size $[1,256,64,64]$.

\subsection{Architectural Adaptations for Wonder3D}
\label{app:wonder3d_architecture}
Wonder3D uses the same decoder stages and channel widths, but their
spatial resolutions are $16\times16$, $32\times32$, and $32\times32$.
Unlike Era3D, each stage is first resized to $32\times32$ and then
processed by a residual bottleneck block. Its convolutional path
uses $1\times1$, $3\times3$, and $1\times1$ convolutions with widths
$C_l\to96\to96\to384$, GroupNorm, and ReLU. A projected shortcut
is added before the final ReLU. Softmax-normalised stage weights
combine these outputs into 384-channel features per domain.

For spatial targets, concatenating the two domains gives
$[3,768,32,32]$ features. The depth mapping uses a $3\times3$
convolution, GroupNorm, SiLU, $2\times$ bilinear upsampling, and a
$1\times1$ projection to produce $[3,256,64,64]$ predictions.
The spatial DINO mapping instead retains $32\times32$ resolution
and projects to 1024 channels. The Wonder3D Caption--T5 mapping
uses colour-domain features from all six views, spatial and view
averaging, and an MLP with widths $384\to768\to768$ and GELU.
Each backbone has its own trained GPAMs; transferring a guidance
configuration does not transfer GPAM parameters.

\subsection{Training and Inference Details}
\label{app:objective_details}
Training and guidance retain the same global or spatial objective
structure. Caption--T5 uses the cosine objective in
Eq.~(\ref{eq:semantic_loss}). Spatial targets use front-view
standardised MSE and two auxiliary $1-\mathrm{CKA}$ terms, each
weighted by $1/2$, as in Eq.~(\ref{eq:geometric_loss}).
During training, standardisation is performed separately for each
channel over spatial locations. Spatial DINO guidance retains this
convention. For Era3D depth guidance, the single-modality and
depth--Caption--T5 configurations instead standardise over the whole
feature map, whereas the depth--DINOv3 configuration uses channel-wise
standardisation. Wonder3D spatial guidance uses channel-wise
standardisation in both training and inference.

The pooled DINO variant spatially pools the target and averages
the pooled predictions across views, then applies the cosine
objective in Eq.~(\ref{eq:semantic_loss}). Its comparison with the
spatial configuration is reported in Appendix~\ref{app:gpam_configuration}.

\label{app:implementation_details}
For Era3D, we train GPAMs using AdamW with learning rate $10^{-4}$, weight decay
0.01, gradient clipping at 1.0, and gradient accumulation over four
steps, with a budget of 10{,}000 updates and validation-based checkpoint
selection. The Wonder3D Caption--T5 GPAM uses 5{,}000 updates without
gradient accumulation. Training takes approximately 1h50m for
Wonder3D and 3h26m for Era3D.
We retain the backbone-specific baseline sampling configurations:
Era3D uses 40 DDIM steps with a classifier-free guidance scale of 3.0,
whereas Wonder3D uses 50 steps with a scale of 1.0.
For each backbone, these settings are held fixed across the baseline,
single-modality, and combined-guidance evaluations, rather than
retuned for OPERA.

For each training object, we render a front-facing RGB image at
$512\times512$ for Era3D or $256\times256$ for Wonder3D.
The Era3D training-pair construction is described in
Appendix~\ref{app:gpam_learning}.

At inference, Era3D inputs use a tight foreground crop: the foreground bounding
box is padded and scaled so that its longest extent spans 420 pixels
within a $512\times512$ canvas. Wonder3D uses an extent of 192 pixels
within a $256\times256$ canvas.

\label{app:latent_updates}
Table~\ref{tab:latent_updates} specifies the update rules used in the
main comparisons. Every guided configuration applies one latent update
at each denoising timestep, before re-evaluating the frozen denoiser
and advancing the scheduler. The guidance scales remain constant
throughout sampling. Unguided baselines perform no latent optimisation.

For combined guidance, each modality's gradient is RMS-normalised
separately over all elements of each domain's latent tensor before
mixing, as in Eq.~(\ref{eq:multi_modality_update}); the RMS denominator
is clamped below at $10^{-8}$. Era3D Caption--T5 single-modality guidance
uses the same normalisation with one objective. The remaining
single-modality configurations use AdamW on the latents with zero
weight decay. This optimiser is reinitialised at every denoising
timestep and takes one step, so it carries no moment estimates between
timesteps. For a gradient component $g$, its update is proportional to
$g/(|g|+\epsilon)$, approximately a sign update, rather than the
domain-wise RMS-normalised update used for combination.

\begin{table}[!htbp]
\centering
\small
\setlength{\tabcolsep}{4pt}
\renewcommand{\arraystretch}{1.05}
\caption{Latent-update configurations in the main comparisons.
All guided configurations use one update per denoising timestep.
The final column specifies standardisation of the front-view MSE,
not gradient normalisation. $H,W$ denotes per-channel spatial
standardisation; $C,H,W$ denotes whole-map standardisation.
A dash denotes the cosine objective, which has no MSE term.}
\label{tab:latent_updates}
\fitresultstable{%
\begin{tabular}{@{}llll@{}}
\toprule
Backbone & Perception used & Latent update & \shortstack[l]{Front MSE\\standardisation} \\
\midrule
Era3D & Depth only & AdamW & $C,H,W$ \\
 & Caption--T5 only & RMS-normalised & --- \\
 & Spatial DINO only & AdamW & $H,W$ \\
 & Depth + Caption--T5 (GSO) & Per-modality RMS & Depth: $C,H,W$ \\
 & Depth + DINOv3 (OmniObject3D) & Per-modality RMS & Both: $H,W$ \\
\midrule
Wonder3D & Depth or spatial DINO only & AdamW & $H,W$ \\
 & Caption--T5 only & AdamW & --- \\
 & Combined (both datasets) & Per-modality RMS & Spatial targets: $H,W$ \\
\bottomrule
\end{tabular}%
}
\end{table}

Thus, comparisons between combined and single-modality configurations
include differences in the latent-update rule and, for some depth
configurations, feature standardisation. They compare the reported
settings rather than holding every optimisation choice fixed while
adding a second modality.

\clearpage
\section{Paired Statistical Analysis}
\label{app:paired_statistics}
Table~\ref{tab:paired_baseline_summary} compares combined perceptual guidance
against the original baselines on GSO and OmniObject3D, retaining all six evaluation metrics.
Appendix~\ref{app:paired_statistics_details} provides the complete
comparisons with the baseline and each single modality.

\paragraph{Procedure.}
Objects are paired across configurations, with $\Delta>0$ denoting
improvement under combined guidance (the sign is reversed for CD
and LPIPS). We report mean differences, 95\% percentile bootstrap
intervals, and Holm-adjusted two-sided Wilcoxon $p$-values.
The adjustment retains the original family of 18 tests per
backbone--dataset pair: three comparisons across six metrics.

\paragraph{Findings.}
On Wonder3D--GSO, CD, IoU, F1@0.1, and LPIPS comparisons with the
baseline yield $p_{\mathrm{Holm}}<0.05$; CD improves on 24 of 30 objects.
On Era3D--OmniObject3D, PSNR and LPIPS meet this threshold, whereas
the geometry metrics do not.
On Wonder3D--OmniObject3D, IoU, PSNR, and LPIPS meet this threshold,
whereas CD, F1@0.1, and SSIM do not.
On Era3D--GSO, mean CD improves, but the median reduction is
$0.1292\times10^{-3}$, 17 of 30 objects improve, and none of the
six baseline comparisons meets the adjusted threshold.
The results indicate that the consistency and type of benefit
vary across configurations.

\paragraph{Scope.}
These analyses are exploratory: settings were selected on the
evaluation sets, and Holm adjustment does not account for that search.
All arms use one sampling seed (42), so these intervals describe
object-level variation, not reproducibility across sampling runs.

\begin{table}[!htbp]
\centering
\caption{Combined guidance versus the unguided baseline.
Positive mean differences favour combined guidance; intervals are
95\% bootstrap intervals for the mean difference.
The Holm correction includes all 18 tests in each backbone--dataset
family, including the single-modality comparisons reported in
Appendix~\ref{app:paired_statistics_details}.}
\label{tab:paired_baseline_summary}
\small
\setlength{\tabcolsep}{7pt}
\renewcommand{\arraystretch}{1.05}
\fitresultstable{%
\begin{tabular}{@{}lrrr@{}}
\toprule
Metric & Mean $\Delta$ & 95\% CI & $p_{\mathrm{Holm}}$ \\
\midrule
\multicolumn{4}{l}{\textit{Era3D--GSO ($n=30$)}} \\
CD ($\times10^{-3}$) & $+5.0116$ & $[+0.2474,\,+11.3350]$ & $1.000$ \\
IoU & $+0.0435$ & $[-0.0014,\,+0.1021]$ & $1.000$ \\
F1@0.1 & $+0.0407$ & $[+0.0021,\,+0.0904]$ & $1.000$ \\
PSNR & $+0.8868$ & $[+0.0764,\,+1.9834]$ & $1.000$ \\
SSIM & $+0.0061$ & $[+0.0001,\,+0.0142]$ & $1.000$ \\
LPIPS & $+0.0092$ & $[+0.0003,\,+0.0215]$ & $1.000$ \\
\midrule
\multicolumn{4}{l}{\textit{Era3D--OmniObject3D ($n=100$)}} \\
CD ($\times10^{-3}$) & $+8.8870$ & $[+3.0607,\,+15.0851]$ & $0.166$ \\
IoU & $+0.0125$ & $[-0.0205,\,+0.0455]$ & $1.000$ \\
F1@0.1 & $+0.0224$ & $[-0.0168,\,+0.0623]$ & $1.000$ \\
PSNR & $+0.8042$ & $[+0.3501,\,+1.2602]$ & $0.006$ \\
SSIM & $+0.0043$ & $[+0.0008,\,+0.0077]$ & $0.089$ \\
LPIPS & $+0.0096$ & $[+0.0049,\,+0.0143]$ & $0.003$ \\
\midrule
\multicolumn{4}{l}{\textit{Wonder3D--GSO ($n=30$)}} \\
CD ($\times10^{-3}$) & $+1.1838$ & $[+0.3903,\,+2.1500]$ & $0.005$ \\
IoU & $+0.0102$ & $[+0.0030,\,+0.0176]$ & $0.039$ \\
F1@0.1 & $+0.0151$ & $[+0.0062,\,+0.0259]$ & $0.001$ \\
PSNR & $+0.2125$ & $[+0.0592,\,+0.4012]$ & $0.060$ \\
SSIM & $+0.0014$ & $[-1.2\!\times\!10^{-5},\,+0.0032]$ & $0.192$ \\
LPIPS & $+0.0022$ & $[+0.0006,\,+0.0044]$ & $0.039$ \\
\midrule
\multicolumn{4}{l}{\textit{Wonder3D--OmniObject3D ($n=100$)}} \\
CD ($\times10^{-3}$) & $+0.1092$ & $[-1.7799,\,+1.8099]$ & $0.320$ \\
IoU & $+0.0182$ & $[+0.0043,\,+0.0335]$ & $0.015$ \\
F1@0.1 & $+0.0117$ & $[-0.0031,\,+0.0272]$ & $0.345$ \\
PSNR & $+0.3520$ & $[+0.0986,\,+0.6291]$ & $0.010$ \\
SSIM & $+0.0018$ & $[+0.0006,\,+0.0031]$ & $0.077$ \\
LPIPS & $+0.0021$ & $[+0.0007,\,+0.0037]$ & $0.004$ \\
\bottomrule
\end{tabular}%
}
\end{table}

\section{Semantic Perception Ablation}
\label{app:semantic_ablation}

\subsection{Guidance Scale}
\label{app:guidance_step_size}
Table~\ref{tab:learning_rate} compares guidance scale $\eta_s$ settings for Caption--T5, DINOv2, and DINOv3 on Era3D--GSO.
For all three targets, $\eta_s=10^{-3}$ yields the lowest CD among the
tested values; larger updates do not improve this metric.
We therefore use $\eta_s=10^{-3}$ for semantic guidance.
This is an inference step size, not the learning rate used to train
GPAMs. Its reuse on OmniObject3D is a shared-setting choice,
not a claim that it is optimal there.


\begin{table}[!htbp]
\centering
\small
\setlength{\tabcolsep}{3pt}
\renewcommand{\arraystretch}{0.95}
\caption{Semantic guidance-step-size comparison on Era3D--GSO.}
\label{tab:learning_rate}
\fitresultstable{%
\begin{tabular}{@{}lrrrrrr@{}}
\toprule
$\eta_s$ & CD$\downarrow$ & IoU$\uparrow$ & F1@.1$\uparrow$ & PSNR$\uparrow$ & SSIM$\uparrow$ & LPIPS$\downarrow$ \\
\midrule
\multicolumn{7}{l}{\textit{Caption--T5}} \\
$10^{-3}$ & \textbf{15.51} & \textbf{0.529} & \textbf{0.803} & \textbf{21.70} & \textbf{0.9235} & \textbf{0.0690} \\
$2.5\times10^{-3}$ & 18.93 & 0.514 & 0.783 & 20.94 & 0.9182 & 0.0764 \\
$10^{-2}$ & 20.06 & 0.468 & 0.781 & 20.28 & 0.9141 & 0.0870 \\
\midrule
\multicolumn{7}{l}{\textit{DINOv2}} \\
$10^{-3}$ & \textbf{15.13} & \textbf{0.534} & \textbf{0.800} & \textbf{21.43} & \textbf{0.9213} & \textbf{0.0695} \\
$2.5\times10^{-3}$ & 16.68 & 0.526 & 0.782 & 21.35 & 0.9208 & 0.0706 \\
$10^{-2}$ & 17.39 & 0.526 & 0.788 & 21.41 & 0.9195 & 0.0718 \\
\midrule
\multicolumn{7}{l}{\textit{DINOv3}} \\
$10^{-3}$ & \textbf{16.28} & \textbf{0.538} & \textbf{0.796} & 21.31 & \textbf{0.9203} & \textbf{0.0714} \\
$2.5\times10^{-3}$ & 16.76 & 0.528 & 0.787 & 21.26 & 0.9201 & 0.0716 \\
$10^{-2}$ & 18.05 & 0.526 & 0.773 & \textbf{21.41} & 0.9200 & 0.0719 \\
\bottomrule
\end{tabular}%
}
\end{table}


\subsection{GPAM with DINOv3}
\label{app:gpam_configuration}
Table~\ref{tab:gpam_configuration} compares three DINOv3 configurations
on Era3D--OmniObject3D with $\eta_s=10^{-3}$. The pooled configuration
averages patch-token features into a global target and applies
a cosine objective to predictions pooled across all six views.
The spatial configuration retains the feature maps, combining
direct front-view alignment with auxiliary CKA losses at two
neighbouring views. Auxiliary CKA provides a relaxed structural
comparison, allowing neighbouring-view predictions to differ
from the input target in their individual feature values.

\paragraph{Impact of CKA.}
To assess the auxiliary terms directly, we retain the same trained
spatial GPAM and remove only the two CKA terms during inference,
leaving front-view standardised MSE as the guidance objective.
The head still reads views $\{0,1,5\}$; the checkpoint, guidance
scale, guided domains, AdamW update, sampling settings, and seed
are unchanged. This comparison tests inference-time auxiliary
guidance, not the effect of CKA during GPAM training.

Spatial + CKA achieves the lowest mean CD and the best PSNR, SSIM,
and LPIPS among the three configurations. Compared with front-only
guidance, it reduces mean CD from 29.03 to 28.89 and improves PSNR
by approximately 0.09\,dB and LPIPS by 0.0007. The front-only variant
has higher IoU and F1@0.1, while pooling gives the highest values
on these two metrics overall. Paired tests between the two spatial
variants do not establish a CD improvement ($p=0.40$); PSNR and
LPIPS favour auxiliary CKA with unadjusted $p=0.014$ and $p=0.044$,
respectively. These exploratory tests use one sampling seed and
are not corrected for multiple comparisons.
We retain Spatial + CKA for its mean CD and appearance results;
the controlled comparison suggests a modest appearance benefit,
rather than a demonstrated geometric advantage from the auxiliary
terms. The pooled comparison additionally changes the perception mapping and
supervised views, so it does not isolate the effect of spatial
representation alone.

\begin{table}[!htbp]
\centering
\small
\setlength{\tabcolsep}{3pt}
\renewcommand{\arraystretch}{0.95}
\caption{Performance of our method with different DINOv3 configurations
and alignment objectives on Era3D--OmniObject3D
(100 objects, $\eta_s=10^{-3}$).
The two spatial variants use the same trained GPAM and differ only
in whether auxiliary CKA is applied at inference. The pooled variant
also changes the perception mapping, supervised views, and objective.
Values are means.}
\label{tab:gpam_configuration}
\fitresultstable{%
\begin{tabular}{@{}lrrrrrr@{}}
\toprule
Variant & CD$\downarrow$ & IoU$\uparrow$ & F1@.1$\uparrow$ & PSNR$\uparrow$ & SSIM$\uparrow$ & LPIPS$\downarrow$ \\
\midrule
Spatial + CKA & \textbf{28.89} & 0.503 & 0.693 & \textbf{21.02} & \textbf{0.9183} & \textbf{0.0773} \\
Spatial, front only & 29.03 & 0.505 & 0.698 & 20.93 & 0.9179 & 0.0780 \\
Pooled & 29.83 & \textbf{0.508} & \textbf{0.701} & 20.89 & 0.9179 & 0.0777 \\
\bottomrule
\end{tabular}%
}
\end{table}

\section{Depth Perception Ablation}
\label{app:depth_ablation}

\subsection{Depth Representation}
\label{app:depth_representation}
Table~\ref{tab:depth_representation} compares intermediate depth features
with the output depth map at three guidance scales $\eta_d$.
At $\eta_d=2.5\times10^{-3}$, intermediate features outperform the
depth-map target on all six metrics, reducing CD from 12.87 to
11.64. This result supports using intermediate perceptual features
rather than the final depth prediction for the selected configuration.
The ordering is not uniform across guidance scales: at $10^{-2}$, the
depth-map target has lower CD and higher F1@0.1. We thus select the
representation together with its guidance strength, rather than
claiming that intermediate features are always superior.


\begin{table}[!htbp]
\centering
\small
\setlength{\tabcolsep}{3pt}
\renewcommand{\arraystretch}{0.95}
\caption{Performance of our method with different depth representations on Era3D--GSO. Best values per metric are selected across both representations.}
\label{tab:depth_representation}
\fitresultstable{%
\begin{tabular}{@{}lrrrrrr@{}}
\toprule
$\eta_d$ & CD$\downarrow$ & IoU$\uparrow$ & F1@.1$\uparrow$ & PSNR$\uparrow$ & SSIM$\uparrow$ & LPIPS$\downarrow$ \\
\midrule
\multicolumn{7}{l}{\textit{Intermediate depth features}} \\
$10^{-3}$ & 12.15 & 0.564 & 0.831 & \textbf{22.23} & \textbf{0.9273} & \textbf{0.0621} \\
$2.5\times10^{-3}$ & \textbf{11.64} & 0.558 & \textbf{0.833} & 22.21 & 0.9272 & 0.0624 \\
$10^{-2}$ & 13.69 & \textbf{0.570} & 0.820 & 21.80 & 0.9252 & 0.0665 \\
\midrule
\multicolumn{7}{l}{\textit{Depth map}} \\
$10^{-3}$ & 12.93 & 0.553 & 0.826 & 22.05 & 0.9263 & 0.0637 \\
$2.5\times10^{-3}$ & 12.87 & 0.548 & 0.826 & 22.08 & 0.9266 & 0.0635 \\
$10^{-2}$ & 13.04 & 0.556 & 0.826 & 21.86 & 0.9250 & 0.0674 \\
\bottomrule
\end{tabular}%
}
\end{table}

\subsection{Guidance Scale}
\label{app:depth_guidance_step_size}
For intermediate depth features, the sweep in
Table~\ref{tab:depth_representation} gives the lowest CD and highest
F1@.1 at $\eta_d=2.5\times10^{-3}$, while $10^{-3}$ gives slightly
better appearance scores. We select $\eta_d=2.5\times10^{-3}$
to favour CD. As with semantic guidance, its reuse on OmniObject3D
does not establish optimality on that dataset.

\subsection{CKA for Depth-Map Targets}
\label{app:depth_cka}
Linear CKA is invariant to orthogonal transformations of the feature
coordinates and isotropic scaling~\citep{DBLP:conf/icml/Kornblith0LH19}.
It therefore provides a relaxed comparison rather than a guarantee
of geometric correspondence. For non-constant single-channel maps,
it reduces to squared Pearson correlation, up to numerical
stabilisation, and is insensitive to correlation sign.
This limitation is relevant to the depth-map ablation:
the auxiliary CKA terms cannot distinguish positive from negative
correlation, while the front-view standardised MSE remains
sensitive to sign inversions.

\clearpage
\section{More Qualitative Results}
\label{app:more_qualitative}
\begin{figure}[!htbp]
\centering
\includegraphics[width=\linewidth,height=0.72\textheight,keepaspectratio]{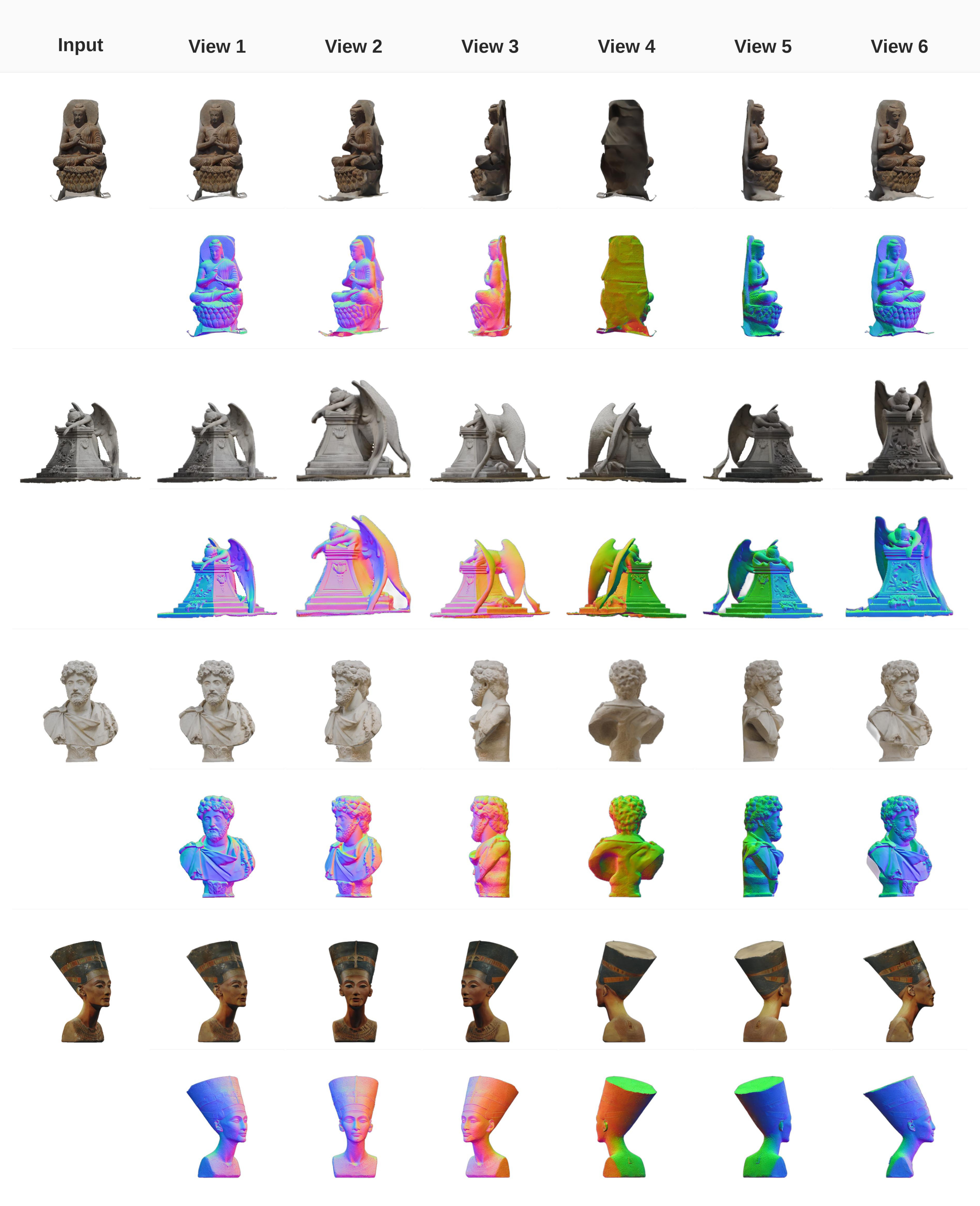}
\caption{\textbf{Multi-view generation on Objaverse.} Era3D with combined semantic and depth guidance. For each object, the first column shows the input image; the remaining columns show six generated colour views (upper row) and normal maps (lower row).}
\label{fig:multiview_objaverse}
\end{figure}

\begin{figure}[!htbp]
\centering
\includegraphics[width=\linewidth,height=0.78\textheight,keepaspectratio]{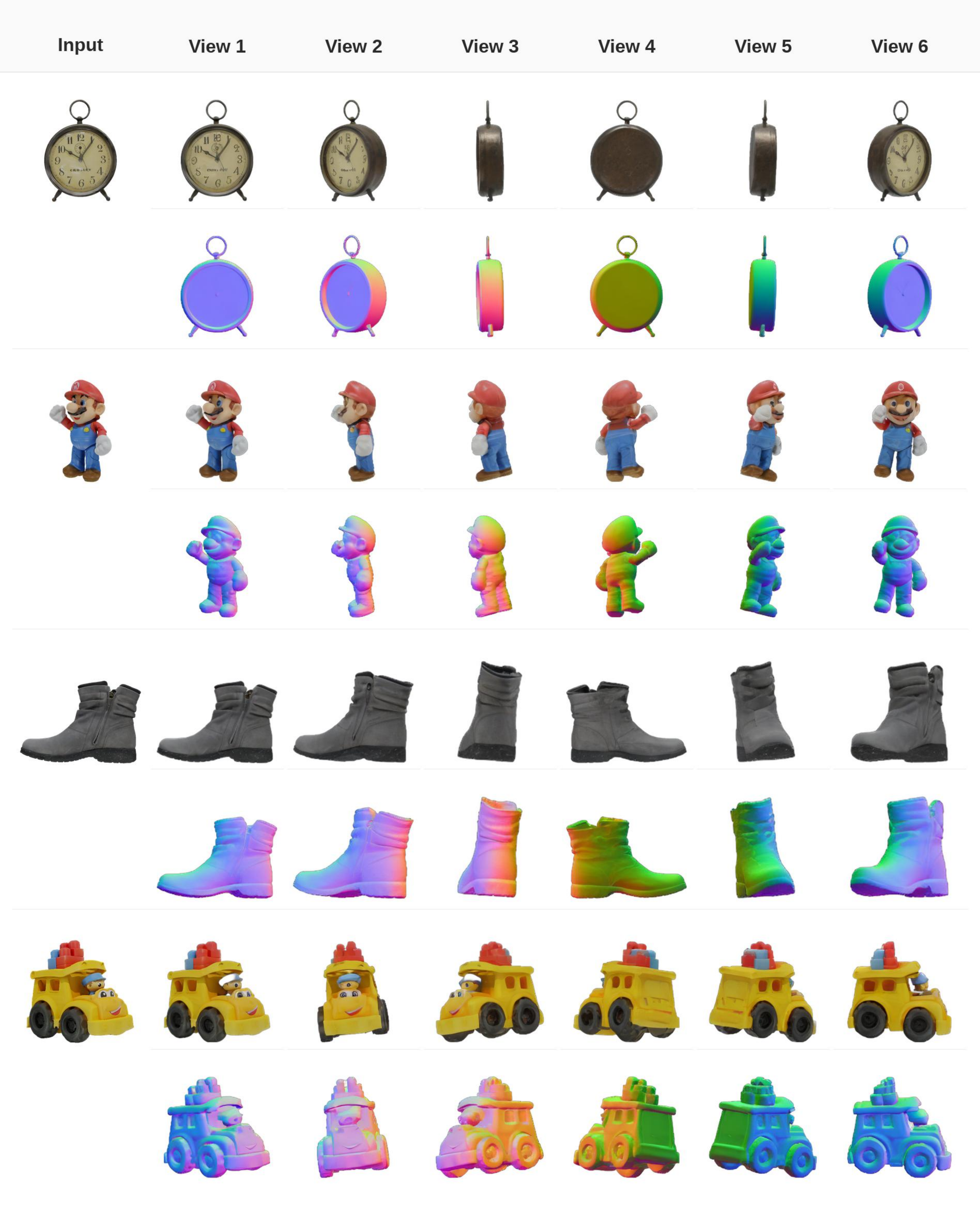}
\caption{\textbf{Multi-view generation on GSO.} Wonder3D with combined semantic and depth guidance. For each object, the first column shows the input image; the remaining columns show six generated colour views (upper row) and normal maps (lower row).}
\label{fig:multiview_gso}
\end{figure}

\begin{figure}[!htbp]
\centering
\includegraphics[width=\linewidth,height=0.78\textheight,keepaspectratio]{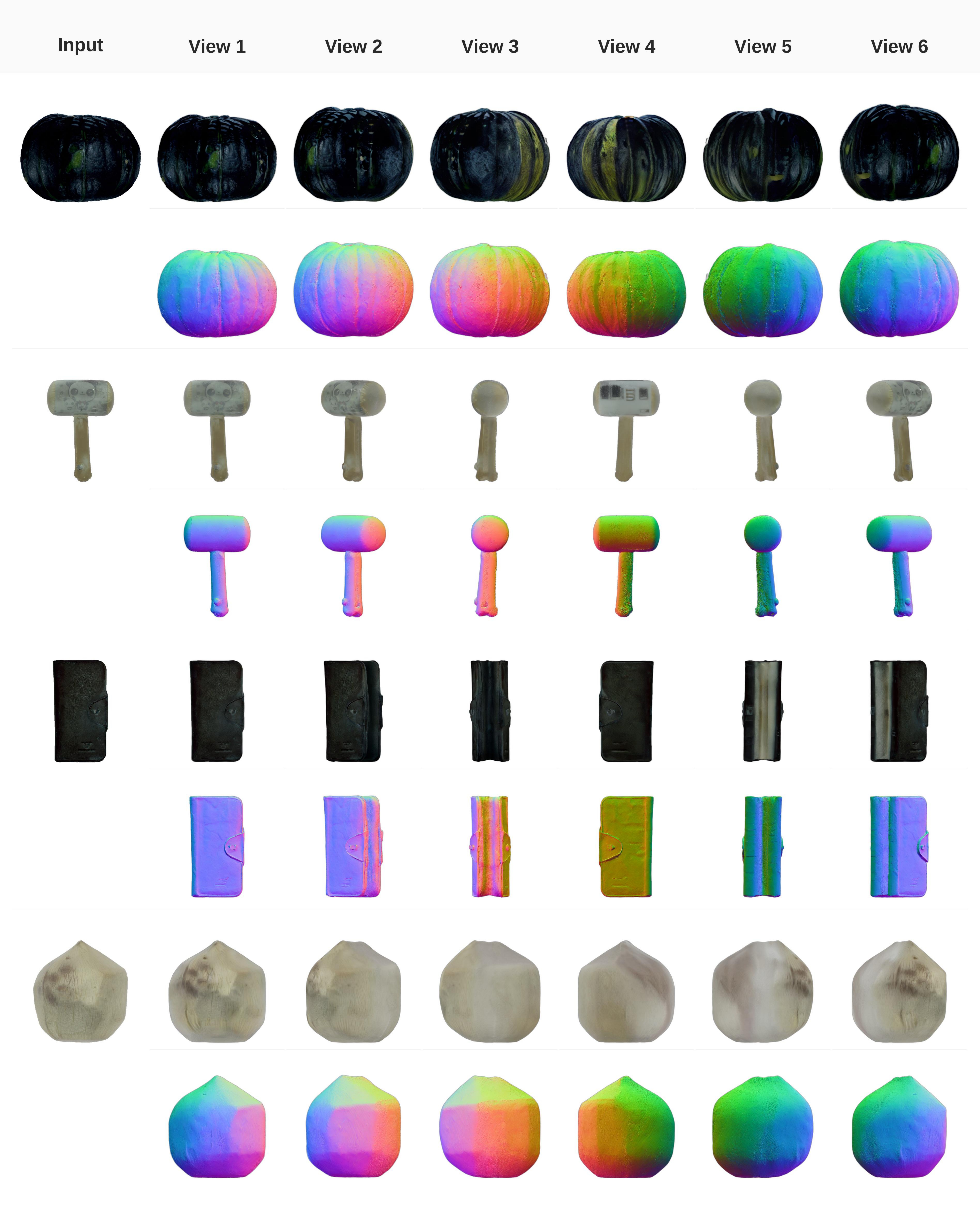}
\caption{\textbf{Multi-view generation on OmniObject3D.} Era3D with combined semantic and depth guidance. For each object, the first column shows the input image; the remaining columns show six generated colour views (upper row) and normal maps (lower row).}
\label{fig:multiview_omni}
\end{figure}

We provide additional multi-view generation results with combined semantic
and depth guidance on Objaverse, GSO, and OmniObject3D
(Figures~\ref{fig:multiview_objaverse}--\ref{fig:multiview_omni}).
Each example shows the input image and six generated colour views with
their corresponding normal maps, before surface reconstruction.

\clearpage
\section{Comparison with External Image-to-3D Generators}
\label{app:external_comparison}
Table~\ref{tab:external_generators} compares selected OPERA
configurations with SAM~3D Objects~\citep{sam3dteam2025sam3d3dfyimages}
and TRELLIS.2~\citep{Xiang_2026_CVPR}.
We use Era3D with OPERA on GSO and Wonder3D with OPERA on
OmniObject3D.
The evaluated OPERA configurations achieve higher mean PSNR and
SSIM and lower mean LPIPS, while the external generators obtain
higher F1@0.1.
These results highlight a remaining appearance-fidelity gap in the
evaluated image-to-3D generators under our protocol: stronger geometric
scores do not necessarily coincide with more faithful rendered appearance.
The selected qualitative examples further illustrate alterations to
object-specific details despite plausible overall shapes.
This gap is not an inevitable trade-off; the goal is to recover
geometrically accurate objects while faithfully preserving their
appearance from the available visual evidence.
These comparisons evaluate complete pipelines; the contribution
of guidance is measured against each corresponding backbone.

\begin{table}[!htbp]
\centering
\caption{Comparison with external generators on GSO (30 objects) and
OmniObject3D (100 objects). Values are means.
OPERA uses combined depth and semantic guidance.
F1@0.1 uses a distance threshold of 0.1 unit-sphere radii.
Appearance scores depend on the pose handling described below.}
\label{tab:external_generators}
\small
\setlength{\tabcolsep}{8pt}
\renewcommand{\arraystretch}{1.1}
\fitresultstable{%
\begin{tabular}{@{}lrrrr@{}}
\toprule
Method & F1@0.1$\uparrow$ & PSNR$\uparrow$ & SSIM$\uparrow$ & LPIPS$\downarrow$ \\
\midrule
\multicolumn{5}{l}{\textit{GSO}} \\
Era3D + OPERA & 0.842 & \textbf{22.42} & \textbf{0.9283} & \textbf{0.0613} \\
SAM 3D Objects & 0.903 & 18.95 & 0.9180 & 0.0758 \\
TRELLIS.2 & \textbf{0.936} & 19.99 & 0.9213 & 0.0694 \\
\midrule
\multicolumn{5}{l}{\textit{OmniObject3D}} \\
Wonder3D + OPERA & 0.776 & \textbf{22.73} & \textbf{0.9335} & \textbf{0.0649} \\
SAM 3D Objects & \textbf{0.828} & 18.88 & 0.9214 & 0.0764 \\
TRELLIS.2 & 0.825 & 19.37 & 0.9202 & 0.0805 \\
\bottomrule
\end{tabular}%
}
\end{table}

\paragraph{Evaluation protocol.}
We evaluate the same objects and ground-truth meshes.
Geometry is normalised to a unit sphere and aligned by multi-start ICP.
Appearance is evaluated over 16 fixed views at $256\times256$ resolution,
using full-frame images composited on a white background.
SAM~3D uses its predicted input-relative pose. TRELLIS.2 produces a
canonical-frame object; its rendering yaw is selected from 72 candidates
by matching the input image, without ground-truth views.
Consequently, appearance scores reflect pose alignment as well as
geometry and texture, and should not be interpreted as an isolated
comparison of texture quality.

\begin{figure}[p]
\centering
\includegraphics[width=\linewidth,height=0.80\textheight,keepaspectratio]{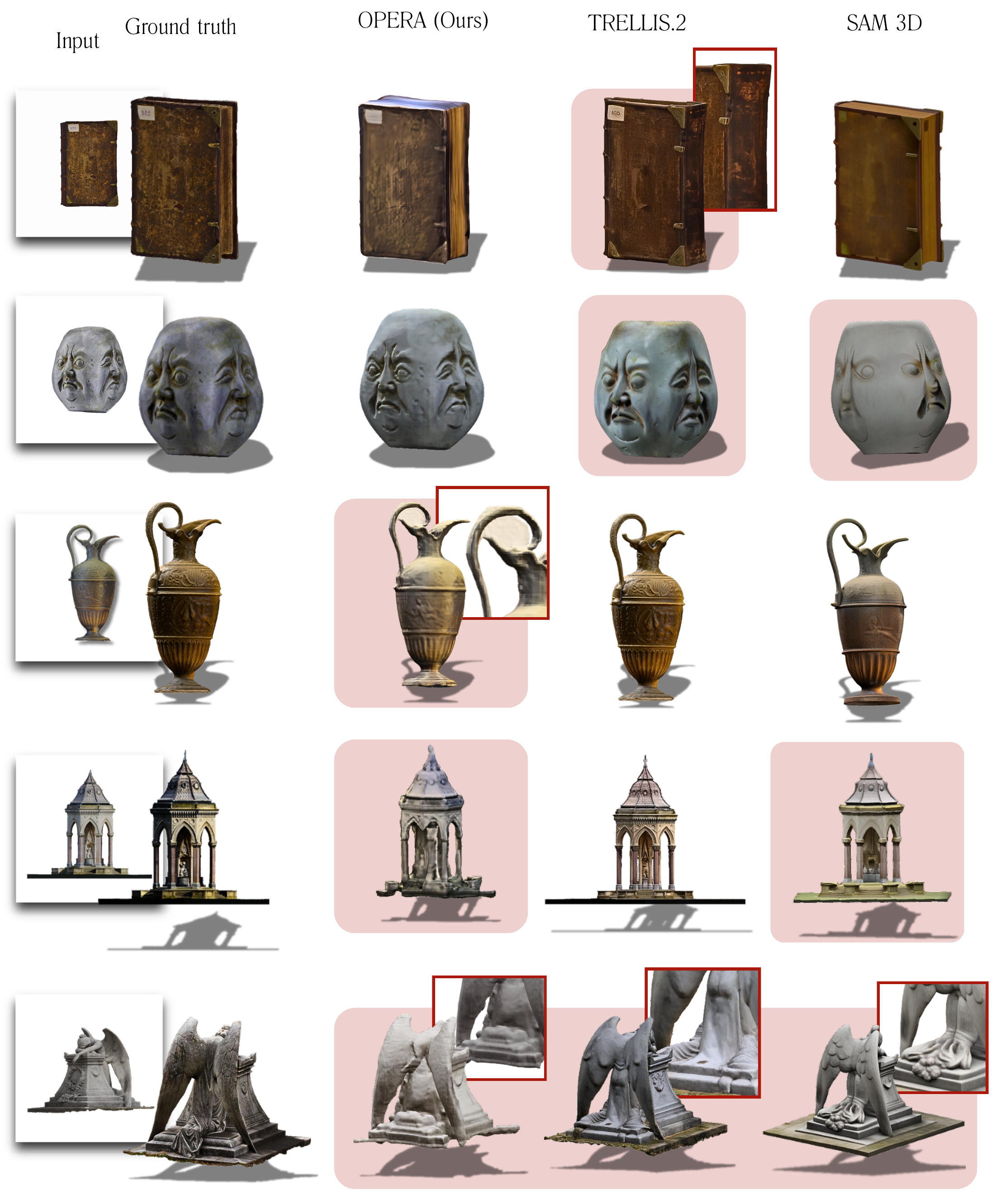}
\caption{\textbf{Failure cases and object-specific fidelity.}
From left to right: input image, ground truth, OPERA, TRELLIS.2,
and SAM~3D Objects. Red shading and red-bordered close-ups highlight
selected errors, including surface discontinuities, degraded details,
and altered object-specific features.}
\label{fig:external_failure_cases}
\end{figure}

\paragraph{Failure cases and object-specific fidelity.}
The selected examples in Figure~\ref{fig:external_failure_cases}
reveal limitations of OPERA in fine-detail fidelity and surface
integrity, including local discontinuities in the vase handle,
degraded architectural details, and fragmented sculptural surfaces.
Despite these defects, the reconstructions remain recognisable and
broadly preserve the reference objects' characteristic forms.
The external references illustrate a complementary challenge:
sharper or more elaborate details do not necessarily correspond to
greater fidelity. TRELLIS.2 introduces decorative patterns on the
book that differ from the ground truth and alters the facial forms
of the four-faced sculpture, while SAM~3D substantially reduces
the sculpture's facial definition.
These examples distinguish local degradation of detail from the
introduction or alteration of object-defining features.
They show that OPERA outputs can retain object-specific form while
remaining limited in fine-scale accuracy and surface continuity.
These are selected pipeline-level observations, not an isolated
assessment of the effect of guidance or a general performance ranking.

\clearpage
\section{Evaluation Subset}
\label{app:evaluation_subset}
\begin{figure}[htbp]
\centering
\includegraphics[width=\linewidth,height=0.78\textheight,keepaspectratio]{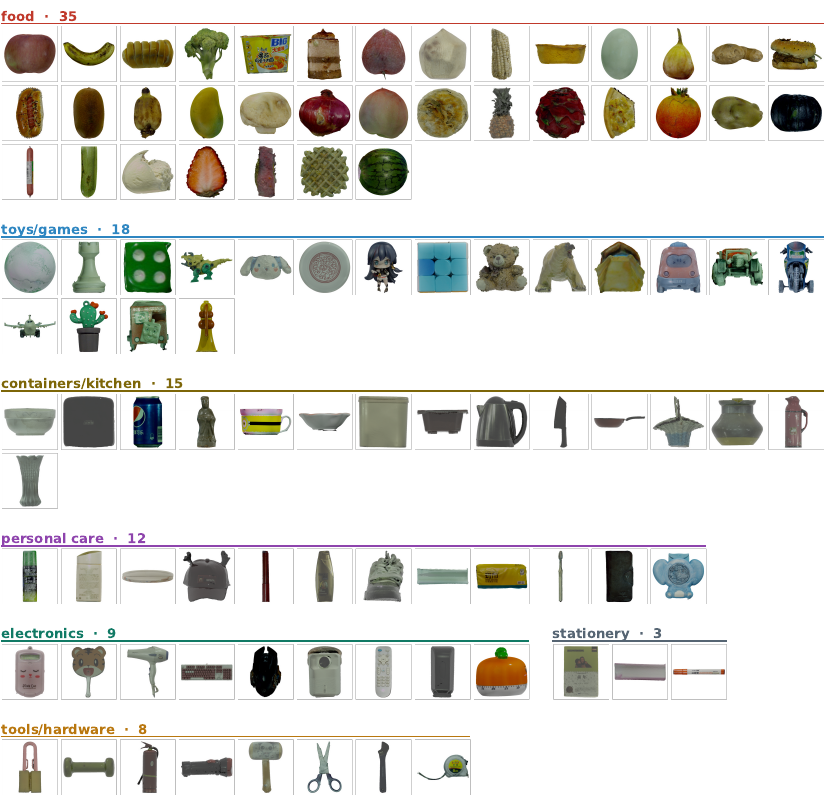}
\caption{\textbf{OmniObject3D evaluation subset.} Input images of all 100 evaluated objects, arranged into seven broad groups for visualisation; group labels and counts describe this subset, not the full dataset.}
\label{fig:benchmark_composition}
\end{figure}

We evaluate on 100 objects from OmniObject3D spanning diverse object
categories (Figure~\ref{fig:benchmark_composition}).

\clearpage
\section{Detailed Paired Comparisons}
\label{app:paired_statistics_details}
Tables~\ref{tab:paired_e3_gso}--\ref{tab:paired_w3_omni} report all
comparisons underlying Appendix~\ref{app:paired_statistics},
including combined guidance versus each single modality.
All six metrics are retained for every comparison.

\paragraph{Statistical procedure.}
We pair configurations by object identity without excluding objects
and orient differences so that $\Delta>0$ favours combined guidance.
We report mean and median paired differences, counts of improved
objects, and 95\% percentile bootstrap intervals from 20,000 object
resamples (bootstrap seed 20260921).
Two-sided Wilcoxon signed-rank tests exclude zero differences from
ranking. Raw and Holm-adjusted $p$-values are reported, with 18 tests
per backbone--dataset family.
Mean-difference intervals and signed-rank tests need not agree:
the former estimate a mean effect, while the latter use the ranks
and signs of paired differences.

\paragraph{Single-modality comparisons.}
The combined and single-modality configurations differ in their
update rules and, in some cases, feature standardisation.
These comparisons therefore assess the reported configurations
rather than isolating the causal effect of combining modalities.
A nonsignificant result does not establish equivalence.

\begin{table}[!htbp]
\centering
\caption{Paired comparisons for Era3D on GSO ($n=30$). Positive differences favour combined guidance. Intervals describe the mean paired difference; $p_{\mathrm{Holm}}$ adjusts the 18 tests in this table.}
\label{tab:paired_e3_gso}
\small
\setlength{\tabcolsep}{4pt}
\renewcommand{\arraystretch}{1.05}
\fitresultstable{%
\begin{tabular}{@{}lrrrrrr@{}}
\toprule
Metric & Mean $\Delta$ & Median $\Delta$ & 95\% CI & Improved & $p$ & $p_{\mathrm{Holm}}$ \\
\midrule
\multicolumn{7}{l}{\textit{Combined vs baseline}} \\
CD ($\times10^{-3}$) & $+5.0116$ & $+0.1292$ & $[+0.2474,\,+11.3350]$ & 17/30 & $0.280$ & $1.000$ \\
IoU & $+0.0435$ & $+0.0059$ & $[-0.0014,\,+0.1021]$ & 18/30 & $0.213$ & $1.000$ \\
F1@0.1 & $+0.0407$ & $+0.0039$ & $[+0.0021,\,+0.0904]$ & 17/30 & $0.140$ & $1.000$ \\
PSNR & $+0.8868$ & $+0.0828$ & $[+0.0764,\,+1.9834]$ & 18/30 & $0.135$ & $1.000$ \\
SSIM & $+0.0061$ & $-4.2\!\times\!10^{-6}$ & $[+0.0001,\,+0.0142]$ & 15/30 & $0.253$ & $1.000$ \\
LPIPS & $+0.0092$ & $+0.0010$ & $[+0.0003,\,+0.0215]$ & 16/30 & $0.119$ & $1.000$ \\
\midrule
\multicolumn{7}{l}{\textit{Combined vs depth-only}} \\
CD ($\times10^{-3}$) & $+0.1851$ & $+0.0181$ & $[-0.3913,\,+0.8934]$ & 18/30 & $0.824$ & $1.000$ \\
IoU & $+0.0167$ & $+0.0027$ & $[-0.0013,\,+0.0460]$ & 18/30 & $0.164$ & $1.000$ \\
F1@0.1 & $+0.0086$ & $+0.0002$ & $[-0.0024,\,+0.0224]$ & 15/30 & $0.670$ & $1.000$ \\
PSNR & $+0.2086$ & $+0.0098$ & $[-0.0342,\,+0.5797]$ & 15/30 & $0.516$ & $1.000$ \\
SSIM & $+0.0011$ & $+0.0002$ & $[-0.0002,\,+0.0030]$ & 17/30 & $0.503$ & $1.000$ \\
LPIPS & $+0.0010$ & $+0.0001$ & $[-0.0006,\,+0.0034]$ & 15/30 & $0.919$ & $1.000$ \\
\midrule
\multicolumn{7}{l}{\textit{Combined vs semantic-only}} \\
CD ($\times10^{-3}$) & $+4.0624$ & $-0.0070$ & $[+0.0806,\,+10.3066]$ & 15/30 & $0.280$ & $1.000$ \\
IoU & $+0.0453$ & $+0.0025$ & $[+0.0013,\,+0.1061]$ & 17/30 & $0.191$ & $1.000$ \\
F1@0.1 & $+0.0389$ & $+0.0076$ & $[+0.0026,\,+0.0852]$ & 19/30 & $0.084$ & $1.000$ \\
PSNR & $+0.7175$ & $+0.2532$ & $[-0.0391,\,+1.7020]$ & 21/30 & $0.058$ & $1.000$ \\
SSIM & $+0.0048$ & $+0.0007$ & $[-0.0003,\,+0.0118]$ & 18/30 & $0.237$ & $1.000$ \\
LPIPS & $+0.0076$ & $+0.0017$ & $[2.5\!\times\!10^{-5},\,+0.0183]$ & 20/30 & $0.092$ & $1.000$ \\
\bottomrule
\end{tabular}%
}
\end{table}

\begin{table}[!htbp]
\centering
\caption{Paired comparisons for Era3D on OmniObject3D ($n=100$). Positive differences favour combined guidance. Intervals describe the mean paired difference; $p_{\mathrm{Holm}}$ adjusts the 18 tests in this table.}
\label{tab:paired_e3_omni}
\small
\setlength{\tabcolsep}{4pt}
\renewcommand{\arraystretch}{1.05}
\fitresultstable{%
\begin{tabular}{@{}lrrrrrr@{}}
\toprule
Metric & Mean $\Delta$ & Median $\Delta$ & 95\% CI & Improved & $p$ & $p_{\mathrm{Holm}}$ \\
\midrule
\multicolumn{7}{l}{\textit{Combined vs baseline}} \\
CD ($\times10^{-3}$) & $+8.8870$ & $+3.1500$ & $[+3.0607,\,+15.0851]$ & 58/100 & $0.017$ & $0.166$ \\
IoU & $+0.0125$ & $+0.0041$ & $[-0.0205,\,+0.0455]$ & 53/100 & $0.522$ & $1.000$ \\
F1@0.1 & $+0.0224$ & $2.4\!\times\!10^{-5}$ & $[-0.0168,\,+0.0623]$ & 50/100 & $0.487$ & $1.000$ \\
PSNR & $+0.8042$ & $+0.7042$ & $[+0.3501,\,+1.2602]$ & 67/100 & $4.0\!\times\!10^{-4}$ & $0.006$ \\
SSIM & $+0.0043$ & $+0.0043$ & $[+0.0008,\,+0.0077]$ & 60/100 & $0.007$ & $0.089$ \\
LPIPS & $+0.0096$ & $+0.0087$ & $[+0.0049,\,+0.0143]$ & 68/100 & $1.5\!\times\!10^{-4}$ & $0.003$ \\
\midrule
\multicolumn{7}{l}{\textit{Combined vs depth-only}} \\
CD ($\times10^{-3}$) & $+6.6046$ & $+0.9274$ & $[+0.9801,\,+12.7775]$ & 53/100 & $0.116$ & $0.929$ \\
IoU & $-0.0142$ & $-0.0036$ & $[-0.0432,\,+0.0141]$ & 47/100 & $0.464$ & $1.000$ \\
F1@0.1 & $+0.0069$ & $-0.0010$ & $[-0.0306,\,+0.0446]$ & 47/100 & $0.667$ & $1.000$ \\
PSNR & $+0.5359$ & $+0.4215$ & $[+0.1397,\,+0.9366]$ & 63/100 & $0.012$ & $0.129$ \\
SSIM & $+0.0026$ & $+0.0032$ & $[-0.0003,\,+0.0053]$ & 56/100 & $0.072$ & $0.644$ \\
LPIPS & $+0.0058$ & $+0.0072$ & $[+0.0020,\,+0.0097]$ & 64/100 & $0.007$ & $0.089$ \\
\midrule
\multicolumn{7}{l}{\textit{Combined vs semantic-only}} \\
CD ($\times10^{-3}$) & $+1.9986$ & $+0.2225$ & $[-0.5759,\,+4.7711]$ & 53/100 & $0.483$ & $1.000$ \\
IoU & $+0.0090$ & $+0.0040$ & $[-0.0128,\,+0.0303]$ & 60/100 & $0.128$ & $0.929$ \\
F1@0.1 & $+0.0120$ & $+0.0070$ & $[-0.0093,\,+0.0330]$ & 57/100 & $0.153$ & $0.929$ \\
PSNR & $+0.2314$ & $+0.2140$ & $[-0.0512,\,+0.5060]$ & 64/100 & $0.002$ & $0.023$ \\
SSIM & $+0.0022$ & $+0.0021$ & $[+0.0005,\,+0.0039]$ & 73/100 & $3.9\!\times\!10^{-4}$ & $0.006$ \\
LPIPS & $+0.0027$ & $+0.0021$ & $[+0.0004,\,+0.0049]$ & 69/100 & $3.2\!\times\!10^{-4}$ & $0.005$ \\
\bottomrule
\end{tabular}%
}
\end{table}

\begin{table}[!htbp]
\centering
\caption{Paired comparisons for Wonder3D on GSO ($n=30$). Positive differences favour combined guidance. Intervals describe the mean paired difference; $p_{\mathrm{Holm}}$ adjusts the 18 tests in this table.}
\label{tab:paired_w3_gso}
\small
\setlength{\tabcolsep}{4pt}
\renewcommand{\arraystretch}{1.05}
\fitresultstable{%
\begin{tabular}{@{}lrrrrrr@{}}
\toprule
Metric & Mean $\Delta$ & Median $\Delta$ & 95\% CI & Improved & $p$ & $p_{\mathrm{Holm}}$ \\
\midrule
\multicolumn{7}{l}{\textit{Combined vs baseline}} \\
CD ($\times10^{-3}$) & $+1.1838$ & $+0.3164$ & $[+0.3903,\,+2.1500]$ & 24/30 & $3.1\!\times\!10^{-4}$ & $0.005$ \\
IoU & $+0.0102$ & $+0.0042$ & $[+0.0030,\,+0.0176]$ & 21/30 & $0.003$ & $0.039$ \\
F1@0.1 & $+0.0151$ & $+0.0033$ & $[+0.0062,\,+0.0259]$ & 24/30 & $6.3\!\times\!10^{-5}$ & $0.001$ \\
PSNR & $+0.2125$ & $+0.0761$ & $[+0.0592,\,+0.4012]$ & 21/30 & $0.010$ & $0.060$ \\
SSIM & $+0.0014$ & $+0.0002$ & $[-1.2\!\times\!10^{-5},\,+0.0032]$ & 21/30 & $0.038$ & $0.192$ \\
LPIPS & $+0.0022$ & $+0.0010$ & $[+0.0006,\,+0.0044]$ & 21/30 & $0.003$ & $0.039$ \\
\midrule
\multicolumn{7}{l}{\textit{Combined vs depth-only}} \\
CD ($\times10^{-3}$) & $+0.7982$ & $+0.1420$ & $[+0.0076,\,+1.9559]$ & 20/30 & $0.058$ & $0.192$ \\
IoU & $+0.0068$ & $+0.0013$ & $[+0.0005,\,+0.0138]$ & 21/30 & $0.096$ & $0.192$ \\
F1@0.1 & $+0.0137$ & $+0.0050$ & $[+0.0042,\,+0.0253]$ & 20/30 & $0.005$ & $0.045$ \\
PSNR & $+0.1645$ & $+0.0341$ & $[+0.0580,\,+0.3017]$ & 23/30 & $0.003$ & $0.033$ \\
SSIM & $+0.0011$ & $+0.0002$ & $[+0.0002,\,+0.0023]$ & 21/30 & $0.047$ & $0.192$ \\
LPIPS & $+0.0019$ & $+0.0006$ & $[+0.0007,\,+0.0034]$ & 25/30 & $3.1\!\times\!10^{-4}$ & $0.005$ \\
\midrule
\multicolumn{7}{l}{\textit{Combined vs semantic-only}} \\
CD ($\times10^{-3}$) & $+1.2056$ & $+0.2959$ & $[+0.3360,\,+2.3436]$ & 22/30 & $4.6\!\times\!10^{-4}$ & $0.007$ \\
IoU & $+0.0098$ & $+0.0034$ & $[+0.0026,\,+0.0176]$ & 23/30 & $0.004$ & $0.043$ \\
F1@0.1 & $+0.0130$ & $+0.0046$ & $[+0.0039,\,+0.0242]$ & 20/30 & $0.007$ & $0.053$ \\
PSNR & $+0.1934$ & $+0.0662$ & $[+0.0545,\,+0.3538]$ & 21/30 & $0.007$ & $0.053$ \\
SSIM & $+0.0011$ & $+0.0001$ & $[-2.7\!\times\!10^{-5},\,+0.0024]$ & 17/30 & $0.105$ & $0.192$ \\
LPIPS & $+0.0019$ & $+0.0009$ & $[+0.0006,\,+0.0035]$ & 24/30 & $0.002$ & $0.033$ \\
\bottomrule
\end{tabular}%
}
\end{table}

\begin{table}[!htbp]
\centering
\caption{Paired comparisons for Wonder3D on OmniObject3D ($n=100$).
Positive differences favour combined guidance.
Intervals describe the mean paired difference;
$p_{\mathrm{Holm}}$ adjusts the 18 tests in this table.}
\label{tab:paired_w3_omni}
\small
\setlength{\tabcolsep}{4pt}
\renewcommand{\arraystretch}{1.05}
\fitresultstable{%
\begin{tabular}{@{}lrrrrrr@{}}
\toprule
Metric & Mean $\Delta$ & Median $\Delta$ & 95\% CI & Improved & $p$ & $p_{\mathrm{Holm}}$ \\
\midrule
\multicolumn{7}{l}{\textit{Combined vs baseline}} \\
CD ($\times10^{-3}$) & $+0.1092$ & $+0.1043$ & $[-1.7799,\,+1.8099]$ & 62/100 & $0.058$ & $0.320$ \\
IoU & $+0.0182$ & $+0.0033$ & $[+0.0043,\,+0.0335]$ & 66/100 & $0.001$ & $0.015$ \\
F1@0.1 & $+0.0117$ & $+0.0003$ & $[-0.0031,\,+0.0272]$ & 53/100 & $0.103$ & $0.345$ \\
PSNR & $+0.3520$ & $+0.0612$ & $[+0.0986,\,+0.6291]$ & 66/100 & $7.3\!\times\!10^{-4}$ & $0.010$ \\
SSIM & $+0.0018$ & $+0.0002$ & $[+0.0006,\,+0.0031]$ & 61/100 & $0.009$ & $0.077$ \\
LPIPS & $+0.0021$ & $+0.0005$ & $[+0.0007,\,+0.0037]$ & 66/100 & $2.3\!\times\!10^{-4}$ & $0.004$ \\
\midrule
\multicolumn{7}{l}{\textit{Combined vs depth-only}} \\
CD ($\times10^{-3}$) & $+0.1105$ & $+0.0377$ & $[-0.6259,\,+0.8234]$ & 60/100 & $0.086$ & $0.345$ \\
IoU & $+0.0049$ & $+0.0013$ & $[-0.0011,\,+0.0110]$ & 59/100 & $0.013$ & $0.105$ \\
F1@0.1 & $-0.0025$ & $+0.0000$ & $[-0.0105,\,+0.0052]$ & 45/100 & $0.865$ & $0.865$ \\
PSNR & $+0.2137$ & $+0.0430$ & $[+0.1010,\,+0.3580]$ & 69/100 & $2.2\!\times\!10^{-5}$ & $4.0\!\times\!10^{-4}$ \\
SSIM & $+0.0009$ & $+0.0003$ & $[+0.0004,\,+0.0016]$ & 63/100 & $4.2\!\times\!10^{-4}$ & $0.006$ \\
LPIPS & $+0.0011$ & $+0.0004$ & $[+0.0005,\,+0.0018]$ & 66/100 & $3.1\!\times\!10^{-5}$ & $5.2\!\times\!10^{-4}$ \\
\midrule
\multicolumn{7}{l}{\textit{Combined vs semantic-only}} \\
CD ($\times10^{-3}$) & $+0.0852$ & $+0.1681$ & $[-1.6172,\,+1.5135]$ & 62/100 & $0.053$ & $0.320$ \\
IoU & $+0.0129$ & $+0.0028$ & $[+0.0012,\,+0.0259]$ & 64/100 & $0.004$ & $0.041$ \\
F1@0.1 & $+0.0059$ & $+0.0003$ & $[-0.0062,\,+0.0177]$ & 53/100 & $0.131$ & $0.345$ \\
PSNR & $+0.2847$ & $+0.0405$ & $[+0.0455,\,+0.5360]$ & 60/100 & $0.004$ & $0.040$ \\
SSIM & $+0.0013$ & $+0.0001$ & $[+0.0002,\,+0.0025]$ & 58/100 & $0.040$ & $0.283$ \\
LPIPS & $+0.0015$ & $+0.0004$ & $[+0.0002,\,+0.0030]$ & 67/100 & $0.002$ & $0.021$ \\
\bottomrule
\end{tabular}%
}
\end{table}

\clearpage

\end{document}